\pdfoutput=1
\documentclass{article}
\usepackage{microtype}
\usepackage{graphicx}
\usepackage{booktabs}
\usepackage{hyperref}

\usepackage[preprint]{icml2026}
\usepackage{amsmath,amssymb}


\icmltitlerunning{Blind to What They Can Fix: A Detection Gap in Self-Improving Agents}
\begin{document}
\twocolumn[
  \icmltitle{Blind to What They Can Fix:\\A Detection Gap in Self-Improving Agents}
  \icmlsetsymbol{equal}{*}
  \begin{icmlauthorlist}
    \icmlauthor{Teng Cao}{equal,aiml,hessian,cs}
    \icmlauthor{Yu Deng}{equal,aiml,hessian,cs}
    \icmlauthor{Kristian Kersting}{aiml,hessian,cs,dfki,cogsci}
  \end{icmlauthorlist}
  \icmlaffiliation{aiml}{Artificial Intelligence and Machine Learning Lab, Technical University of Darmstadt, Germany}
  \icmlaffiliation{hessian}{Hessian Center for Artificial Intelligence (hessian.AI), Germany}
  \icmlaffiliation{cs}{Department of Computer Science, Technical University of Darmstadt, Germany}
  \icmlaffiliation{dfki}{German Research Center for Artificial Intelligence (DFKI), Germany}
  \icmlaffiliation{cogsci}{Centre for Cognitive Science, Technical University of Darmstadt, Germany}
  \icmlcorrespondingauthor{Teng Cao}{teng.cao@tu-darmstadt.de}
  \icmlcorrespondingauthor{Yu Deng}{yu.deng@tu-darmstadt.de}
  \icmlkeywords{recursive self-improvement, programmatic agents, runtime monitoring, autonomous maintenance}
  \vskip 0.3in
]
\printAffiliationsAndNotice{\icmlEqualContribution}

\begin{abstract}
AI agents can write programs that carry out a task without calling a model at every step, so the task runs fast and its logic stays open to inspection. Agents improve these programs by running them, studying their failures, and revising the code, a form of recursive self-improvement (RSI). We ask whether this process also yields runtime checks that can be trusted. In five task domains, with up to seven author models, agents bring their programs to high scores. Yet when the environment changes, the checks they were told to write stay silent on half or more of the resulting failures in four domains. These failures can be detected and repaired: a monitor built from records of normal runs catches most of them in three domains, and the same agents usually repair them once called. The checks do not examine the recorded relations that reveal the failures, and the strongest agent writes checks that catch most of them once told that its checks are the only thing that calls it back. An unreported failure stays unrepaired, which limits what self-improvement is worth in autonomous operation. Evaluations of self-improvement should measure not only scores but also whether a system notices its own failures.
\end{abstract}

\section{Introduction}
\textbf{Programs written by AI agents run tasks efficiently and keep their decision logic in plain view.} Such a program makes recurring decisions without calling a large language model before each action. Its logic is written as explicit conditions, rules, and actions that people can read, test, reuse, and change \citep{zhang2026copitree}.

\begin{figure*}[t]
\centering
\includegraphics[width=\textwidth]{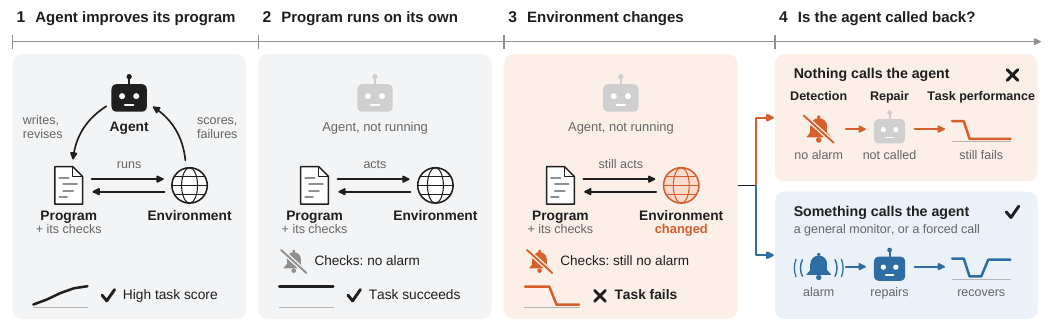}
\caption{\textbf{An agent can repair a failure that its own program never reports.} (1) The agent improves a task program and its runtime checks over rounds of execution feedback. (2) The program then runs without the agent. (3) A change in the environment makes the task fail, while the program keeps acting and its checks keep passing. (4) If nothing calls the agent, the failure persists; if a monitor or the experimenter calls it, the agent repairs the program. We measure detection, repair, and task performance on the same failures. The diagram is schematic, and the small curves in its panels show no measured data.}
\label{fig:loop}
\end{figure*}

\textbf{An agent can improve such a program from its own execution: it runs the program, studies where it failed, revises the code, and tests the revision.} Each round builds on the last, a feedback-driven form of recursive self-improvement (RSI) \citep{wu2026rsigame,cao2026kintsugi}. Once the program runs on its own, with no model watching each step, a new question arises: will the system notice a new failure and start the next round of improvement?

\textbf{A high task score does not tell us whether a system can notice new failures.} During development the agent tests the program and sees every failure. During deployment nobody is watching, so the system itself must decide when something needs attention, and a change in the environment can make a good program fail while every action it takes remains valid. We therefore change regularities of the environment that programs may rely on, confirm that the task can still be done, and measure three things on the same pairs of program and change: task performance, failure detection, and repair when the agent is called.

\textbf{In five task domains, with up to seven author models, agents revise programs to high scores yet do not write checks that reliably detect task failures.} Some program formats ask the agent to state what its rules rely on and to check it at every step. These checks stay silent on half or more of the failures in four of the five domains, and no threshold on how often they are contradicted turns them into a usable alarm. The failures themselves are neither hidden nor hard to fix: a general-purpose monitor built from records of normal runs detects most of them in three domains, and the same agents repair most of the missed failures when called again, without being told what changed.

\textbf{The failures that the checks miss are visible in execution records that the checks never look at.} We keep the program fixed and add checks on recorded relations, that is, on statistics of the recorded variables. Most misses disappear, even when the 32 added relations are drawn at random. Purpose matters as well. Told that a broken check is the only thing that calls it back, Codex closes most of the gap itself: with new programs its misses fall from 69\% to 6\% in Seaquest and from 95\% to 0\% in AppWorld, whereas three open models still miss 84\% to 100\%. The gap is costly. In replayed deployments a trigger that fires in time recovers most of the lost task value, whereas a perfect repairer behind the program's own checks recovers almost none (Figure~\ref{fig:loop}).
\par \textbf{We contribute an evaluation framework, evidence of a detection gap, and an analysis of its causes and its consequences for deployment.}
\begin{enumerate}\setlength{\itemsep}{3pt}\setlength{\parsep}{0pt}\setlength{\parskip}{0pt}
\item \textbf{We introduce a controlled framework for evaluating autonomous maintenance by programmatic agents.} It measures task performance, failure detection, and repair on the same pairs of program and change.
\item \textbf{We identify a detection gap in feedback-driven program improvement.} The checks that agents are told to write miss half or more of the task failures in four of five domains, although the failures can be detected from execution records and the same agents repair most of them when they are called.
\item \textbf{We trace the gap to what the checks cover and measure what it costs.} Adding recorded relations to the checks of a fixed program removes most misses. Codex writes checks that miss few failures once it is told what they are for, while open models still miss most. In replayed deployments, timely calls recover most of the lost value, and a better repairer behind the program's own checks recovers almost none.
\end{enumerate}

\section{Programmatic Agents and Self-Improvement}

\textbf{A programmatic agent writes a program once and lets it act, and recursive self-improvement revises that program from its own execution feedback.} The agent writes a task program, and the program then chooses actions without asking the agent each time \citep{zhang2026copitree}. In each round the agent runs the program, reads its records and failures, revises the code, and tests the revision \citep{wu2026rsigame,cao2026kintsugi}. The agent itself stays fixed; what improves is the program. A version is kept only if it performs the task better (Appendix~\ref{app:protocol}), so nothing in the loop measures or rewards how reliable the program's checks are.

\textbf{A program that runs on its own still needs a way to call the agent back.} Runtime checks are code that inspects observations or execution records and flags possible problems. A monitor is any mechanism that turns such signals into an alarm, and an alarm calls the agent back to investigate what happened and revise the program.

\section{Evaluation Framework}

\begin{figure*}[t]
\centering
\includegraphics[width=\textwidth]{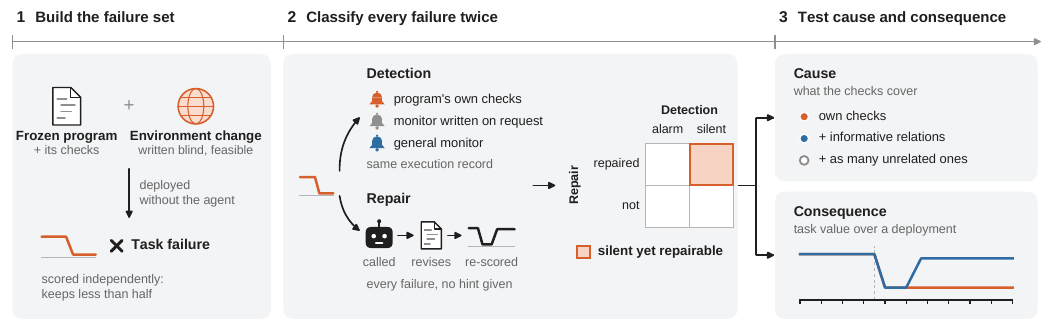}
\caption{\textbf{Every failure is classified twice: by whether it was detected and by whether it can be repaired.} (1) A frozen program and its checks are deployed, without the agent, under an environment change written without seeing the checks; a deployment that keeps less than half of its task performance is a task failure. (2) Detection: the same execution record goes to the program's own checks and to detectors outside the program. Repair: the agent is called on every failure, with no hint about the change, and the revised program is scored again. Failures that raised no alarm yet were repaired form the detection gap. (3) Further experiments test the cause, by changing what the checks cover, and the consequence, by replaying deployments. The curves are schematic.}
\label{fig:framework}
\end{figure*}

\textbf{The framework first builds a set of task failures and then asks the same questions of every failure.} We freeze each program, deploy it under controlled changes to its environment, and identify task failures from outcomes that are scored independently. For every failure we record whether each detector raises an alarm and whether the agent that wrote the program repairs it when called (Figure~\ref{fig:framework}).

\subsection{Building the Failure Set}

\textbf{Controlled changes to the environment test whether programs and their checks still work when conditions shift.} Let $E_0$ be the original environment and $E_\delta=\mathcal{T}_\delta(E_0)$ the environment after an intervention $\delta$ on a regularity that a program may rely on: what an action does, a resource or workload condition, a delay in an observation, or the meaning of an API response. No change used for evaluation is seen during development. All but nine were written by isolated sessions that saw neither the programs under test nor their results; we call these blind changes. A change is kept only if the task can still be done, as shown by a reference solution or by independent successful runs. Deployments without a change, and changes under which a program keeps working, serve as controls for needless alarms.

\textbf{We decide what counts as a task failure from independently scored outcomes, before any detector is compared.} A unit $u=(\pi,\delta)$ is one frozen program under one change, run for ten deployment episodes (days, in the service and API domains). Let $J_u^0$ and $J_u^-$ be its mean task scores in the original and the changed environment, measured by an independent evaluator and without repair. The unit is a task failure when the program keeps less than half of its original performance in the changed environment:
\begin{equation}
F_u=\mathbf{1}\{J_u^-<\theta_u\},\qquad \theta_u=\tfrac{1}{2}J_u^0.
\label{eq:failure}
\end{equation}
All detectors are judged on the same failure set $\mathcal{F}=\{u:F_u=1\}$. The results change little with thresholds of 0.3 or 0.7, and no Seaquest program drops below the threshold between two sets of unchanged episodes (Appendix~\ref{app:detection}).

\subsection{Detection and Repair on the Same Failures}

\textbf{Every detector reads the same execution records, and we count its misses, its false alarms, and its alarms on programs that still work.} A detector sees the records of a unit's episodes and is never told whether or when the environment changed. An episode counts as alarmed if the detector alarms at any step. A detector $M$ is silent on a failure, $d_u^M=0$, if it alarms in fewer than half of the unit's episodes, and its miss rate is the share of failures on which it is silent. Alarms on unchanged episodes are false alarms. We report separately the alarms on changed units in which the program keeps working, where a call is just as unnecessary as after a false alarm.

\textbf{To measure repair, we call the same agent on every failure, including the failures that raised no alarm.} The agent $A$ that wrote the program gets one session per failed unit, with a fixed budget of tool calls. It may run the program in the changed environment, on episodes other than those used for scoring, and revise it. It is not told what changed. Let $J_u^A$ be the score of the revised program under the same evaluator. The failure is repaired when
\begin{equation}
r_u^A=\mathbf{1}\{J_u^A\geq\theta_u\},
\label{eq:recovery}
\end{equation}
and the share of all failures that a monitor misses but the authoring agent then repairs is
\begin{equation}
\widehat{G}(M,A)=
\frac{1}{|\mathcal{F}|}\sum_{u\in\mathcal{F}}\bigl(1-d_u^M\bigr)\,r_u^A .
\label{eq:gap}
\end{equation}

\subsection{Tasks, Authors, and Monitors}

\textbf{We apply the framework in five task domains.} Seaquest and Kangaroo are arcade games from JAXtari \citep{delfosse2026jaxtari}, and Craftax is an open-ended survival game \citep{matthews2024craftax}. In service autoscaling, the program adds and removes instances of a simulated service so that every hour stays within its targets. In AppWorld, an independently maintained suite of application APIs \citep{trivedi2024appworld}, the program compiles a daily report, and a day succeeds only if the whole report is correct. Seaquest has two catalogues of changes, with 25 and 8 entries; every other domain has one catalogue of 12 (Appendix~\ref{app:catalogues}).

\textbf{Agents built on seven models write the programs.} The main results use Codex (gpt-6.1-sol). Claude Code (opus-5-5) and five open models (Kimi-K2.6, Qwen3.8-27B, Qwen3.8-Flash, Mistral-Medium-3.5, and gemma-4-31B) repeat parts of the study, mainly in service autoscaling and AppWorld (Table~\ref{tab:populations}). Each program is the final version of its own improvement run. The Codex programs complete 1.9 rescues per Seaquest episode and clear two to four Kangaroo levels, where a program before improvement clears none. They meet the service target in about 15 of 16 hours, against 0.7 before improvement and 15.6 for a hand-written threshold policy, and they fill every field of the AppWorld report on days without a change.

\textbf{Some program formats ask the agent to write checks, and these are the program's own checks.} In these formats the instructions say that a rule relies on ``what its actions cost and achieve, on how fast things move, on what stays unaffected''. Each such condition is to be written as an assumption with a check that runs at every step, and the instructions add that an assumption ``that is never contradicted in a changed game tells you nothing''. At each step a check returns a verdict: confirmed, contradicted, or none. A check is broken once it has been contradicted at least three times and more often than confirmed, and an episode with a broken check counts as alarmed. Outside AppWorld, the standard instructions do not say that the program will be deployed or that an alarm calls the author back (Appendix~\ref{app:instructions}). Programs in the other formats contain no checks.

\textbf{Table~\ref{tab:detection} lists the programs and failures behind each detection result.} Repair is measured on the same failures in every domain but Seaquest. There, the 40 programs were deployed under a second catalogue of blind changes, and both detection and repair are measured on its 107 failures. AppWorld needs a note. The two failure criteria that we had registered, that is, fixed in writing before the data existed, produced too few failures. The criterion used here counts a day as failed unless the whole report is correct and treats all 12 changes as feasible. We fixed it after seeing those results and then registered it for a replication with 16 new Codex programs (details in Appendix~\ref{app:detection}).

\textbf{We compare the own checks with three detectors from outside the program.} The general monitor is a single frozen procedure that names no mechanism of any domain. From the records of 20 normal reference episodes per program, it learns the usual ranges of generic statistics of the recorded variables and, where the domain exposes a success event, a floor on the success count. It alarms when a deployment leaves these ranges. Its rules were fixed on Seaquest, after three earlier rounds of changes and before the changes used here, and are applied unchanged in the other domains. The outcome rule is simpler: it alarms when an episode's success count falls below half of the program's reference mean. A monitor on request is written by the author, or by a fresh session of the same model, when we ask for one after learning. Intervals are 95\% bootstrap intervals over programs or, where stated, over programs and changes jointly. Tables~\ref{tab:prereg} to~\ref{tab:prereg4} list every registered prediction with its outcome: of 120 predictions, 69 held, 36 failed, 14 were undetermined, and one held for four of five author models.

\section{Experimental Results}

\textbf{The results answer four scientific questions, each with one experiment on the failures defined above.} (Q1) Do agents that improve programs to high scores also write checks that detect task failures? (Q2) Can the same agent repair the failures that its checks miss? (Q3) Why do the checks miss these failures? (Q4) What does the gap cost during deployment, and is detection or repair the bottleneck?

\subsection{High Scores Do Not Come with Checks~That~Detect~Failures}

\textbf{(Q1) Agents reach high scores without writing checks that reliably notice when those scores are lost.} In deployment, only the program's checks can report a failure, so a silent check leaves a failing program running. We give the same execution records to the program's own checks and to the three detectors from outside the program, and count the failures on which each stays silent.

\begin{table*}[t]
\caption{\textbf{Under the standard instructions, the program's own checks miss half or more of the failures in four domains, for every group of author models.} $P$: programs with checks; $F$: their failures. Misses: failures without an alarm, for the own checks under the format's rule, for the same checks read against 20 reference episodes (Own + ref.), for the outcome rule (not available in AppWorld), and for the general monitor. Brackets: 95\% intervals from resampling programs and changes jointly. New programs are evaluated under the blind changes; informed: Section~\ref{sec:coverage}. $^\dagger$It alarms on 68\% of unchanged Craftax days.}
\label{tab:detection}
\centering
\footnotesize
\setlength{\tabcolsep}{3pt}
\begin{tabular*}{\textwidth}{@{\extracolsep{\fill}}lrrlrrrrrrr@{}}
\toprule
 & & & \multicolumn{4}{c}{Misses on failures (\%)} & \multicolumn{4}{c}{Alarms on working units (\%)} \\
\cmidrule{4-7}\cmidrule{8-11}
Programs with checks & $P$ & $F$ & Own checks & Own + ref. & Outcome & General & Own & Own + ref. & Outcome & General \\
\midrule
Seaquest, Codex & 20 & 168 & 52 [32, 71] & 15 & 7 & 5 & 23 & 23 & 11 & 69 \\
\hspace{0.8em}blind changes only & 20 & 111 & 59 [35, 80] & 17 & 5 & 3 & 23 & 18 & 7 & 67 \\
\hspace{0.8em}new programs & 16 & 90 & 69 [37, 93] & 11 & 2 & 1 & 11 & 33 & 2 & 70 \\
\hspace{0.8em}new programs, informed & 16 & 84 & 6 [0, 21] & 18 & 1 & 4 & 12 & 16 & 5 & 66 \\
Kangaroo, Codex & 6 & 22 & 55 [13, 91] & 18 & 9 & 18 & 8 & 26 & 0 & 58 \\
Service autoscaling, Codex & 4 & 25 & 64 [29, 92] & 20 & 8 & 0 & 13 & 17 & 0 & 52 \\
\hspace{0.8em}six other models & 14 & 82 & 65 [39, 87] & 48 & 5 & 6 & 26 & 10 & 0 & 51 \\
AppWorld, Codex & 4 & 28 & 89 [64, 100] & -- & -- & 61 & 0 & -- & -- & 40 \\
\hspace{0.8em}new programs & 16 & 112 & 95 [84, 100] & -- & -- & 62 & 0 & -- & -- & 40 \\
\hspace{0.8em}new programs, informed & 16 & 122 & 0 [0, 37] & -- & -- & 63 & 9 & -- & -- & 23 \\
\hspace{0.8em}three open models & 12 & 95 & 82 [58, 100] & -- & -- & 59 & 16 & -- & -- & 27 \\
Craftax, Codex & 3 & 8 & 0 & -- & 25 & 0$^\dagger$ & 32 & -- & 7 & 100 \\
\bottomrule
\end{tabular*}
\end{table*}

\begin{figure*}[t]
\centering
\includegraphics[width=\textwidth]{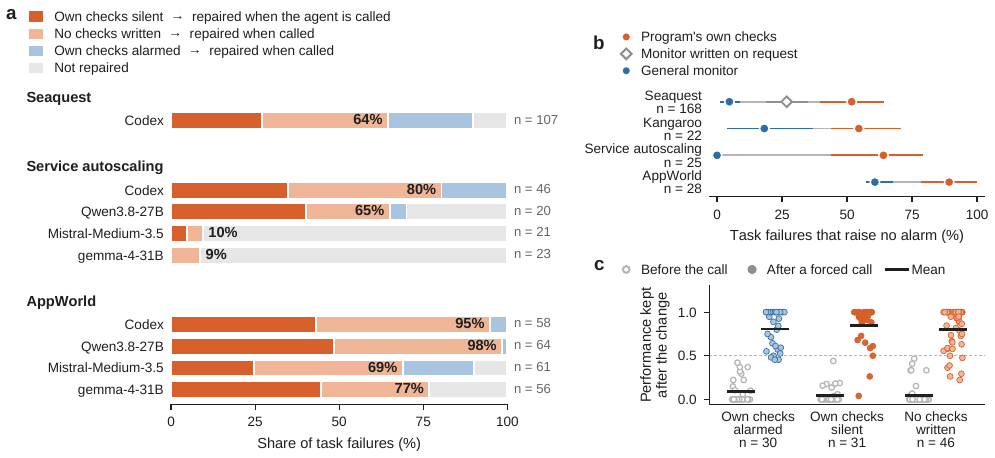}
\caption{\textbf{Most failures that the program's own checks miss are repaired when the agent is called.} (a) Every task failure, by what the program's own checks did and by whether the agent repaired the program when called without being told what had changed; bold numbers give the share of all failures that raised no alarm (checks silent, or no checks written) and were repaired. AppWorld rows use the first study. (b) Failures of Codex programs with checks that raise no alarm: own checks, general monitor and, in Seaquest, the monitor on request. (c) Seaquest, 107 failures: performance kept under the change before and after a forced repair session; the dashed line is the failure threshold. Whiskers are 95\% program-cluster bootstrap intervals.}
\label{fig:uninvoked}
\end{figure*}

\textbf{The program's own checks stay silent on half or more of the failures in four of the five domains, for Codex and for the other author models.} Codex's checks miss 52\% of the failures in Seaquest, 55\% in Kangaroo, 64\% in service autoscaling, and 89\% in AppWorld. The programs of six other models miss 65\% in service autoscaling, and those of three open models miss 82\% in AppWorld (Table~\ref{tab:detection}, Figure~\ref{fig:uninvoked}b). Craftax is the exception: there the Codex checks catch all eight failures. Blind changes and new programs give no better picture. In Seaquest the checks miss 59\% of the failures under the blind changes, against 39\% under the nine changes we wrote ourselves. Sixteen new Codex programs miss 69\% under the blind Seaquest changes, and 16 new AppWorld programs miss 95\% in a registered replication. The failures are concentrated, however: five changes account for 187 of the 295 Seaquest failures (Appendix~\ref{app:detection}). More improvement helps only in part. Over three successive generations of the service-autoscaling programs, the checks miss 95\%, then 79\%, then 64\% of the failures. A different alarm threshold does not help either: no threshold on how often the checks are contradicted turns them into a usable alarm (Figure~\ref{fig:coverage}c, Appendix~\ref{app:detection}).

\textbf{Detectors outside the program miss far fewer of the same failures, and where the outcome can be observed a simple rule on it is enough.} In Seaquest, Kangaroo, and service autoscaling, the outcome rule misses 7\%, 9\%, and 8\% of the failures. It rarely alarms without cause: on at most 5\% of unchanged episodes and at most 11\% of the units in which the program keeps working (Table~\ref{tab:detection}). The general monitor misses 5\%, 18\%, and 0\%, with false alarms on at most 3\% of unchanged episodes. But it reacts to change of any kind, and it alarms on 52\% to 69\% of the units in which the program keeps working. The monitor on request lies in between: across all 40 Seaquest programs it misses 25\% of the failures and alarms on 8\% of the working units.

\textbf{The failures can be detected, so the gap lies in the checks that agents write alongside their programs.} Where a program can observe its own outcome, the checks written under the standard instructions do worse than a one-line rule on that outcome. Codex can also see the problem afterwards. Given unlabelled deployment records, it says that the environment changed for all 120 failing Seaquest episodes and all 24 failing AppWorld days, and for none of the unchanged ones; it judges the program to be failing in 94\% of the failing Seaquest episodes and in 9\% of those under harmless changes (Appendix~\ref{app:detection}). Two domains are harder. In AppWorld a program cannot observe whether its report is correct, and the general monitor misses 61\% of the failures, because the changes alter formats, identities, and nested structure that it does not record. The checks that Codex writes once it knows their purpose catch all of these failures (Section~\ref{sec:coverage}). In Craftax the general monitor alarms on 68\% of unchanged days and is of no use.

\subsection{Agents Repair Most of the~Failures~Their~Checks~Miss}

\textbf{(Q2) The agent repairs most of the failures that its checks miss once it is called, even when the call tells it nothing.} A silent check costs something only if the agent could have repaired the failure. So we call the authoring agent on every failure, whether or not anything alarmed, tell it nothing about the change, and have an independent evaluator score the revised program again (Figure~\ref{fig:uninvoked}a).

\textbf{Codex repairs nine-tenths or more of the failures on which its program raised no alarm.} It repairs 69 of 77 such failures in Seaquest, all 37 in service autoscaling, and all 55 in AppWorld. Repaired Seaquest programs keep 81\% of their original performance under the change, up from 5\%. Among the failures of programs with checks, the share that is both missed and repaired, $\widehat{G}$, is 48\%, 64\%, and 89\%. If programs without checks are counted as raising no alarm, it is 64\%, 80\%, and 95\% of all 107, 46, and 58 failures. One caveat applies. A change under which no frozen program completes the task is excluded as infeasible before repair, and if the one such Seaquest change counted as 40 unrepaired failures, the 64\% would become 47\%. Repair does not depend on whether the checks had noticed the failure (Figure~\ref{fig:uninvoked}c), nor on whether the call carries a runtime report, a one-bit alert, or nothing at all (Appendix~\ref{app:repair}).

\textbf{The result holds for Codex and Qwen3.8-27B wherever we tested it, and for the two weaker models only in AppWorld.} In service autoscaling, 65\% of the failures of Qwen3.8-27B are missed and then repaired. For Mistral-Medium-3.5 and gemma-4-31B the share is only 10\% and 9\%, because their repairs fail there. In AppWorld their shares are 69\% and 77\% (Appendix~\ref{app:repair}).

\subsection{Missed Failures Are Largely a~Matter~of~Check~Coverage}

\textbf{(Q3) The checks miss these failures largely because they examine too few of the relations that the execution records contain.} An example shows how. One Codex program for Seaquest checks that reaching a diver adds one to the divers it carries. A blind change makes divers swim away from the submarine. The submarine never reaches a diver again, so the check never gives a verdict, no episode is alarmed, and the rescues fall to zero (Appendix~\ref{app:example}). \label{sec:coverage}

\begin{figure*}[t]
\centering
\includegraphics[width=\textwidth]{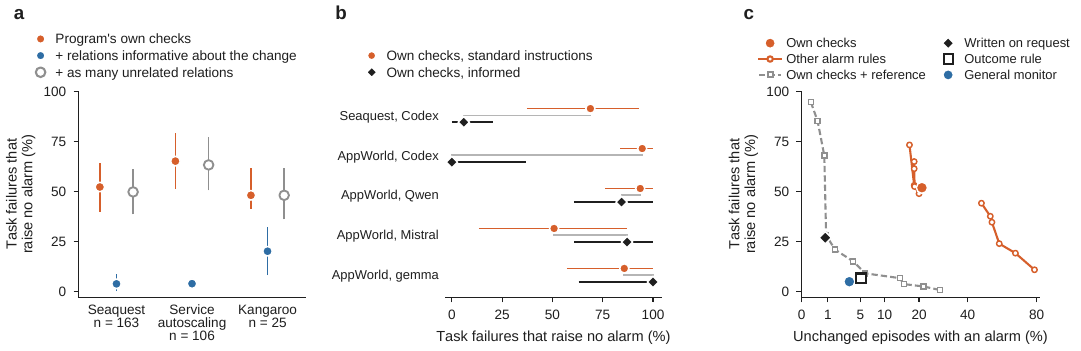}
\caption{\textbf{The checks miss failures that additional recorded relations reveal.} (a) With the task program held fixed: its own checks, alone, with added relations that are informative about the change, and with as many unrelated relations (service autoscaling pools the programs of seven author models; Kangaroo includes one Qwen3.8-Flash program). (b) Failures missed by the own checks of new programs under the standard and the informed instructions, which state the purpose of the checks (Section~\ref{sec:coverage}). (c) Seaquest, 168 failures of the 20 programs with checks: failures missed against unchanged episodes with an alarm (square-root scale), for the own checks under the format's rule, under other alarm rules, and read against reference episodes, and for detectors outside the program. Whiskers are 95\% bootstrap intervals over programs in (a) and over programs and changes in (b).}
\label{fig:coverage}
\end{figure*}

\textbf{We keep the program fixed and change only what its checks cover.} A monitor can be written as $M(h_t)=g(\phi(h_t))$, where $h_t$ is the execution history, $\phi$ selects recorded variables and relations, and $g$ decides whether to alarm. A relation is one statistic of one recorded variable. To the program's own checks we add one of three sets of relations: informative ones, which separate changed from unchanged runs of other programs under the same change and so require knowing the change; equally many unrelated ones, which do not; or relations drawn at random.

\textbf{Informative relations remove most misses, a large enough random set does the same, and unrelated relations remove almost none.} With informative relations, the misses fall from 52\% to 4\% in Seaquest, from 65\% to 4\% in service autoscaling (programs of all seven models), and from 48\% to 20\% in Kangaroo. With the same number of unrelated relations they stay at 50\%, 63\%, and 48\% (Figure~\ref{fig:coverage}a; the comparison uses the 163, 106, and 25 failures that have enough unrelated relations to draw from). False alarms on unchanged runs differ by about two percentage points or less between the conditions. Knowing the change is not necessary: 32 relations drawn at random leave 4\%, 4\%, and 15\% of the failures missed (Appendix~\ref{app:coverage}).

\textbf{A better alarm rule on the checks' verdicts does not recover the failures, but a record of how often the checks apply recovers most of them.} Suppose we learn, from the 20 reference episodes, the usual range of how often each check is contradicted. Alarming outside these ranges still misses 61\% of the Seaquest failures, at a false-alarm rate of 3\%. If the ranges also cover how often each check is confirmed, a quantity that no check examines, the misses fall to 15\% at a false-alarm rate of 4\%. The general monitor misses 5\% at 3\% (Figure~\ref{fig:coverage}c). We chose these settings on these data. Held fixed and applied to programs they were not chosen on, the same rule misses 11\% to 20\% of the failures of Codex programs in Seaquest, Kangaroo, and service autoscaling, and 48\% for the six other models (Table~\ref{tab:detection}).

\textbf{Told that a broken check is the only thing that calls it back, Codex writes checks that miss few failures; the open models still miss most.} We add one paragraph to the instructions that states this purpose (Appendix~\ref{app:e4}) and call the resulting programs informed. In Seaquest, all 16 informed programs add a check that rescues keep happening within a time budget, which no standard program has. Their checks miss 6\% of the failures under the blind changes, against 69\% for the standard programs, and alarm on 10\% of unchanged episodes. Without the new progress checks they would miss 70\%. The alarm rule is the same in both conditions and neither has reference data, so the difference lies in what the checks cover (Figure~\ref{fig:coverage}b). AppWorld gives the same answer by a different route. A program there cannot observe its outcome, and the standard instructions already say that a failed check raises the day's alarm. With the paragraph, the misses of new Codex programs fall from 95\% to 0\%, with no alarm on original days: the informed programs compare the answers of independent API calls. The open models do not follow. With the paragraph, Qwen3.8-27B, Mistral-Medium-3.5, and gemma-4-31B miss 84\%, 87\%, and 100\% of their failures, against 94\%, 51\%, and 86\% without it; the low figure for Mistral arises because half of its standard programs alarm on every day. We had registered the prediction that Codex's misses would persist, and this prediction failed (Tables~\ref{tab:informed} and~\ref{tab:prereg4}).

\textbf{Asking for broad coverage backfires, and examples of failure help only in part.} When we tell the Codex authors of the Seaquest programs to record every kind of object and the range of every number, their monitors miss no failure but alarm on 99.8\% of unchanged episodes. When we instead show an author up to three recorded episodes of its program failing, its misses under blind changes fall from 28\% to 19\%. In the registered analysis, which counts three sessions lost to an infrastructure failure as never alarming, the improvement is 4 points, with an interval from $-4$ to 12 (Appendix~\ref{app:asking}).

\subsection{With a Capable Repairer, Deployment Value Is Limited by Timely Calls}

\textbf{(Q4) Where the agent can repair, what limits the value of a deployment is whether anything calls it in time: the program's own checks often do not, and a perfect repairer behind them adds almost nothing.} A missed failure keeps costing value for as long as the program runs, so the loss has to be measured over a whole deployment.

\begin{figure*}[t]
\centering
\includegraphics[width=\textwidth]{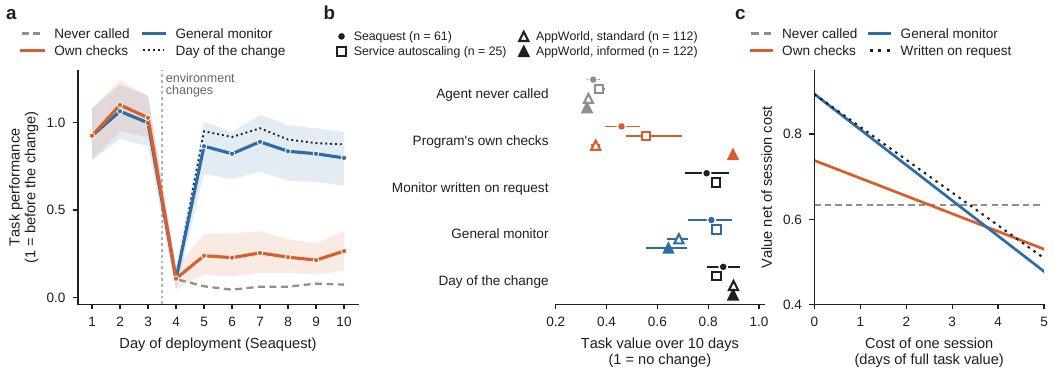}
\caption{\textbf{Where the agent can repair, detection limits the value of the deployment.} (a) Replay of 10-day Seaquest deployments in which the environment changes before day 4 (failing deployments of programs with checks): mean daily task performance relative to before the change, by what calls the agent. (b) Ten-day task value by trigger on such deployments in Seaquest and service autoscaling, and for the new standard and informed AppWorld programs. (c) Service autoscaling, all 48 changed deployments of programs with checks: task value after each session is charged a cost in days of full task value (the lines of the monitor on request and the general monitor nearly coincide). Bands and whiskers are 95\% program-cluster bootstrap intervals.}
\label{fig:value}
\end{figure*}

\textbf{We replay 10-day deployments in which the environment changes before day 4, and vary only what calls the agent.} A trigger opens at most one repair session, after the first day on which it alarms. Every trigger replays the same recorded sessions, so the triggers differ only in when, and whether, they call. Task value is the task success accumulated over the whole deployment, relative to that of the unchanged program in the original environment:
\begin{equation}
V(M,A)=\frac{\mathbb{E}\bigl[\sum_{t=1}^{T}s_t(M,A)\bigr]}{\mathbb{E}\bigl[\sum_{t=1}^{T}s_t^{0}\bigr]}.
\label{eq:deployment-value}
\end{equation}
Here $s_t$ is the task success on day $t$ under trigger $M$ and repairer $A$, and $s_t^0$ is that of the original program without a change. As an upper bound we also consider a perfect repairer, which always restores the original performance.

\textbf{The program's own checks recover a quarter to two-fifths of what the general monitor recovers, and triggers that fire soon after the change recover most of what a perfectly timed call does.} Take the failing deployments of programs with checks, in Seaquest and in service autoscaling. If the agent is never called, the ten-day value is 0.35 and 0.37. The program's own checks raise it to 0.46 and 0.56. The monitor on request reaches 0.79 and 0.83, the general monitor 0.81 and 0.83, and a call on the day of the change 0.86 and 0.83 (Figure~\ref{fig:value}a--b). Swapping the own checks for the general monitor therefore gains 0.35 and 0.28 in value (95\% intervals 0.25--0.45 and 0.15--0.36). Swapping the real repairer for a perfect one, while keeping the own checks, gains 0.01 and 0.03. Part of the Seaquest gain comes from the one-session limit: if calls before the change cost nothing, the gain is 0.21 (0.11--0.32; Appendix~\ref{app:replay}).

\textbf{A timely trigger keeps its advantage when sessions cost something and harmless changes are included.} In service autoscaling every changed deployment has a recorded session, whether or not the program failed, so we can count them all. Over the 48 changed deployments of programs with checks, the value is 0.63 without calls, 0.74 with the own checks, and 0.89 with the general monitor. The general monitor pays for this with more sessions: it opens one in 83\% of deployments, the own checks in 42\%. If each session costs one day of full task value, the two values become 0.70 and 0.81, and on deployments where the program keeps working the general monitor's calls are then a net loss of 0.04. It stays ahead of the own checks until a session costs 3.7 days of task value (Figure~\ref{fig:value}c).

\textbf{Once the purpose is stated, Codex's own checks call it in time.} A forced repair succeeds for all 112 AppWorld failures of the new standard programs and all 122 of the informed ones. In replayed ten-day deployments of the standard programs, the own checks give a value of 0.36, barely above the 0.33 obtained without any call and far below the 0.90 of a call on the day of the change. The own checks of the informed programs reach 0.90. The general monitor gives 0.68 and 0.64 (Figure~\ref{fig:value}b).

\textbf{The trigger can be simple, but it must react to the change, and it helps only as much as the repairer can repair.} A rule that alarms when a day's task success halves does as well as the general monitor, with a value of 0.82 in both domains. For Mistral-Medium-3.5 and gemma-4-31B the general monitor raises value only from 0.38 to 0.46 and 0.47, because their repairs rarely succeed (Appendix~\ref{app:replay}).

\section{Discussion}

\textbf{Improving a program by its score does not, by itself, give the agent a reliable check on that score.} Nothing in the improvement loop tests whether the program would notice a failure. An instruction to check what the rules rely on produces checks on mechanisms, and these leave half or more of the failures unreported in four of the five domains. The ability is there. The same agents repair these failures when called, write far better monitors when asked afterwards, and recognise the change from the records. Codex, once told that its checks are the only call for repair, writes checks that miss few failures; the open models do not. Evaluations of self-improving systems should therefore report detection next to score: the share of induced failures that the system itself reports, together with its alarm rates on unchanged runs and on harmless changes.

\textbf{Noticing a change is easier than noticing a failure, and watching the outcome closes the gap only where the outcome is visible.} A monitor built from normal records notices nearly every change, harmful or not, which is acceptable only when sessions are cheap. A rule on task success, or the progress check of an informed agent, needs an outcome that the program can read at run time. Where there is none, as in AppWorld, informed Codex cross-checks what the service returns. Whether such checks hold up under other changes and over longer deployments remains open.

\subsection{Limitations}
\textbf{The evidence has five limits.} First, the program's own checks exist only in formats that ask for them, and versions are kept by task performance alone. We stated the purpose of the checks only in Seaquest and AppWorld, and we did not test improvement loops that select versions by how well they detect failures. Second, outside Seaquest few independent programs have checks and the catalogues are small. Pooled over all standard programs with checks, the joint intervals of the misses reach down to 40\% in Seaquest, 44\% in service autoscaling, and 8\% in Kangaroo, and only in AppWorld does the lower bound, 73\%, exceed one half (Appendix~\ref{app:detection}); in the first AppWorld study the own checks and the general monitor cannot be told apart under joint resampling (Table~\ref{tab:detection}). Third, the coverage tests rely on knowledge of the change, on the general monitor's vocabulary, or on settings chosen on the evaluation data. They show that the records contain the signal, not that agents find it. Fourth, the deployment results replay recorded sessions with at most one call. Fifth, the general monitor has reference runs that the own checks lack, was developed on Seaquest, reacts to harmless changes, and fails in AppWorld and Craftax.

\section{Related Work}
\textbf{Prior work checks executions with external monitors and judges self-improvement by task performance; we measure whether the improved program reports its own failures.} The oracle problem asks how to tell that an execution is correct \citep{barr2015oracle}. For software, it is addressed by invariant detection, runtime verification, and repair triggered by invariants \citep{ernst2007daikon,leucker2009rv,perkins2009clearview}; for learned systems, by drift detection, data validation, and novelty detection \citep{gama2014drift,breck2019datavalidation,kang2020assertions,boult2021novelty}. Telling a harmful shift from a harmless one has been studied for learned predictors \citep{podkopaev2022harmful,garg2022atc}. Monitors for agents are external components that watch the agent's steps \citep{dubey2026realtime,barazandeh2026ontrack,ji2026sentry,naagarajan2026structural}, and language models can state invariants \citep{pei2023invariants,endres2024nl2postcond} and ask a person for help when they are uncertain \citep{ren2023knowno}.

\textbf{Self-improving systems are judged by the task scores they reach.} Self-improving agents and program search are evaluated in this way \citep{shinn2023reflexion,wang2023voyager,romera2024funsearch,wu2026rsigame,zelikman2024stop,zhang2025dgm}, and agents that keep a world model in code revise it when its predictions fail \citep{tang2024worldcoder}. That our result depends on what the instructions leave open is a case of underspecification \citep{damour2022underspecification}: programs with equally high scores can behave differently once they are deployed.

\textbf{Our protocol and our notion of coverage borrow from software testing.} The protocol is mutation analysis \citep{jia2011mutation} applied to the environment instead of the program, with changes in the manner of HackAtari \citep{delfosse2024hackatari}. What we call coverage has a precedent in checked coverage, the part of a computation that a test's oracles examine \citep{schuler2011checked}. Generated tests serve to select among candidate programs \citep{chen2023codet}, generated oracles are judged by the bugs they find \citep{dinella2022toga}, and the cross-checks that informed Codex writes are metamorphic relations \citep{segura2016metamorphic}. We measure, on the same failures, what the agent's own checks detect, what the agent could repair, and what the difference costs.

\section{Conclusion}
\textbf{Competence and self-maintenance are different properties of an agent system.} Competence is what an agent can accomplish once it is engaged. Self-maintenance depends on whether execution brings the right evidence to the surface and asks for that engagement in time. In five task domains, agents improved their programs to high scores, yet the checks they were told to write stayed silent on half or more of the failures in four of them, although the same agents repaired most of these failures once called. Stating what the checks are for closed most of this gap for Codex but not for the open models. Measuring detection and repair separately, on the same failures, shows how an agent's ability to repair can go unused, and evaluations of self-improving systems should report both measures.

\bibliography{references}
\bibliographystyle{icml2026}

\appendix
\onecolumn
\section{Additional Methods and Results}

\subsection{Program Improvement Protocol}

\textbf{We implement the improvement loop as rounds of execution, revision, and validation.} \label{app:protocol}At round $k$, the retained state is $S_k=(\pi_k,M_k^{\mathrm{self}},N_k)$, comprising the task program, its agent-written runtime checks, and its notes, respectively. Checks may be absent, depending on the program format. The agent configuration $A$ stays fixed across rounds. We first run the current program in the original environment $E_0$ and collect execution feedback $\mathcal{D}_k$:
\begin{equation}
\mathcal{D}_k=\operatorname{Execute}(\pi_k,E_0).
\label{eq:rsi-feedback}
\end{equation}
The feedback records observations, actions, task outcomes, execution errors, and any check outputs, allowing the agent to examine what the program did and where task progress broke down.

\textbf{The agent uses execution feedback to propose and test revisions to the program, its runtime checks, and its notes.} It can run further experiments, modify the task logic or runtime checks, and update its notes within a specified revision budget $b_k$. Let $\mathcal{P}_k$ be the set of candidate versions produced by this process:
\begin{equation}
\mathcal{P}_k=\operatorname{Revise}_A(S_k,\mathcal{D}_k;b_k).
\label{eq:rsi-proposal}
\end{equation}
Each candidate contains a program, its checks, and its notes. Updating the program does not require changing its checks.

\textbf{Validation selects the version that the next round inherits, by task performance alone.} Separate validation runs compare the candidates with the current version by task performance, and the selection rule retains one version:
\begin{equation}
S_{k+1}=\operatorname{Select}_{\mathrm{val}}
\bigl(\{S_k\}\cup\mathcal{P}_k\bigr).
\label{eq:rsi-selection}
\end{equation}
The next round starts from the selected program, checks, and notes. Whether the checks would detect a failure plays no part in the selection. We save the version retained at each round, including rounds without a change. These saved versions, referred to as checkpoints, allow us to evaluate how task performance and detection change during improvement.

\textbf{Relay learning retains a validated version across rounds while allowing several sessions to explore revisions from the same starting point.} Each session records the versions it tests, so an intermediate version can be retained instead of the final edit. Validation ranks these versions by task-completion progress, then task score, with a fixed tie rule. Each session's selected version receives additional validation; the contenders are compared with the version retained from the previous round. The selected program, checks, notes, and an aggregate validation summary are passed to the next round. The random seeds that determine final evaluation runs and the environmental changes used in those runs are withheld from it.

\textbf{The domains implement this loop in two ways, and every evaluated program is the final version of its own run.} In Seaquest and AppWorld a program comes from one learning session: the agent tests versions on learning episodes, is advised to keep a new version only if it is better, and decides when to stop, and the program is the file at the end of the session. Codex sessions take 26 minutes and 44 tool calls at the median in Seaquest, where there is no budget, and 6 minutes and 25 of at most 150 tool calls in AppWorld. In Kangaroo, service autoscaling, and Craftax a program is the champion of a relay of three generations of three parallel sessions with 40 tool calls each, selected on validation episodes as described above. Runs share no state: the 40 Seaquest programs come from 40 separate sessions, ten in each of four formats, and the programs of the other domains from separate sessions or relays (Table~\ref{tab:populations}). Two of the four formats contain checks. In one, every rule states an expectation of what its action should cause and leave unaffected, and the engine checks it and counts the violations; in the other, the program declares assumptions with checks, and a broken assumption also disables the rules that rely on it. The other two formats contain no checks: a free-form program, and a free-form program whose instructions restate the same writing advice in words, with nothing enforced. The other domains use the format with declared assumptions and one format without checks. The generations compared in the main text are successive champions within the service-autoscaling relays. Programs whose mean score in the original environment is below a minimum, 0.5 rescues per Seaquest episode, are excluded before any change is applied to them.

\subsection{Instructions for the Program's Own Checks}

\textbf{The program's own checks are written under an explicit instruction to check what the rules rely on.} \label{app:instructions}In the formats with checks, a program is a set of prioritized rules together with the assumptions about the environment that the rules rely on. Each assumption carries a check that the execution engine runs at every step; once an assumption has been contradicted at least three times and more often than confirmed within an episode, the rules that rely on it are skipped and lower-priority rules take over. The instructions tell the agent that a rule relies on what its actions cost and achieve, on how fast things move, and on what stays unaffected, and that each such condition must be written as an assumption with a check. They also state that an assumption that is never contradicted in a changed environment tells the agent nothing. A program alarms in an episode when one of its assumptions is broken in this sense. The rule is a design choice of the format; a more sensitive rule would trade misses for false alarms, which are already frequent (21\% of unchanged Seaquest episodes). Programs in the other formats are written without declared assumptions and therefore contain no checks.

\textbf{The learning instructions present the checks as part of the policy, and outside AppWorld they do not mention deployment.} The Seaquest instructions contain these sentences: ``Each assumption has a check that the engine runs on every step of play.'' ``When an assumption stops holding during an episode, the rules that rely on it are skipped and lower-priority rules take over.'' ``A rule relies on more than its situation: on what its actions cost and achieve, on how fast things move, on what stays unaffected. Write each such condition as an assumption with a check.'' ``An assumption that is never contradicted in a changed game tells you nothing; a check that returns None when in doubt is better than one that is always True.'' They do not say that the program will be deployed or that a broken assumption calls the author back. The AppWorld instructions do: ``program.py is deployed: it runs every day on its own, and nobody checks the digest'', and ``an assumption that returns False raises an alarm for that day.'' Section~\ref{app:e4} adds this information to the Seaquest instructions.

\subsection{The General Monitor}

\textbf{The general monitor records ranges of generic statistics over normal runs and alarms when a deployment leaves them.} \label{app:monitor}From the object-level observations of a program's normal episodes it computes, over windows of 500 steps, how often each object class moves, appears, or vanishes and how each scalar changes, overall and conditioned on one other variable, together with the values each scalar takes. It keeps the ranges that also hold in held-out normal episodes. It further records discrete regularities, namely the sizes of the changes of each scalar and the intervals between them, and, where the domain exposes a success event, the lowest cumulative success count seen at each point of a normal episode. During deployment it alarms when a statistic leaves its recorded range, when events occur that normal runs never showed, or when the success count falls below its floor. Nothing in the procedure names a mechanism of a particular domain. It uses 20 normal reference episodes per program; the program's own checks have no such reference, which is part of what the comparison measures.

\textbf{The monitor's result does not depend on the success count, and fewer reference episodes mainly raise its false alarms.} Without the success count it misses 6\% of the 295 Seaquest failures, 24\% of the 34 Kangaroo failures, and none of the 46 service-autoscaling failures of Codex programs; the full monitor misses 3\%, 12\%, and 0\%. With 5, 10, or 20 reference episodes in Seaquest it misses 9\%, 2\%, and 3\% and alarms on 9\%, 6\%, and 3\% of unchanged episodes. With ten reference episodes chosen to be alike it alarms on 10\% of unchanged Seaquest episodes and on 18\% in service autoscaling, against 6\% and 1\% for ten episodes that span the range of normal outcomes.

\textbf{The monitor evaluated here is the version fixed before the fourth round of Seaquest changes.} Earlier versions were tested on three rounds of developer-written changes: the first two missed 42\% and 20\% of the failures and were revised, and the third produced too few failures for a verdict (Table~\ref{tab:prereg}). The version fixed after that is used on the 25 changes of rounds four to seven (the first Seaquest catalogue), on the second catalogue, and without modification in the other domains.

\textbf{The general monitor responds to change, whether or not the program fails.} On the 705 changed Seaquest units in which the program keeps at least half of its performance it alarms in 73\% (69\% under the blind changes), against 97\% of the 295 failing units. Its structural statistics alone alarm on 94\% of the failing units, 71\% of the working units, and 2.6\% of unchanged episodes; its success count alone alarms on 65\%, 10\%, and 0.9\%. In Kangaroo, service autoscaling, and AppWorld it alarms on 64\%, 51\%, and 37\% of the working units of Codex programs (Table~\ref{tab:alarms}).

\subsection{Detection: Additional Results}

\begin{table}[h]
\caption{\textbf{Detection by author model.} Share of task failures (\%) that raise no alarm, on the failures of programs that contain their own checks, and for the monitor that the same model wrote when asked (all failures of that model's programs). $P$: programs with checks; $F$: their failures. A dash marks a condition that was not run for that model and domain.}
\label{tab:models}
\centering
\small
\begin{tabular}{@{}llrrrrr@{}}
\toprule
Domain & Author model & $P$ & $F$ & Own checks & General monitor & On request, all failures \\
\midrule
Seaquest & Codex & 20 & 168 & 52 & 5 & 25 \\
Kangaroo & Codex & 6 & 22 & 55 & 18 & -- \\
\midrule
Service autoscaling & Codex & 4 & 25 & 64 & 0 & 0 \\
 & Claude Code & 2 & 12 & 83 & 0 & 5 \\
 & Kimi-K2.6 & 4 & 24 & 38 & 0 & 47 \\
 & Qwen3.8-27B & 2 & 11 & 91 & 9 & 15 \\
 & Mistral-Medium-3.5 & 2 & 11 & 64 & 0 & 62 \\
 & gemma-4-31B & 2 & 12 & 92 & 17 & 17 \\
 & Qwen3.8-Flash & 2 & 12 & 50 & 17 & -- \\
\midrule
AppWorld & Codex & 4 & 28 & 89 & 61 & -- \\
 & Qwen3.8-27B & 4 & 32 & 97 & 53 & -- \\
 & Mistral-Medium-3.5 & 4 & 31 & 48 & 61 & -- \\
 & gemma-4-31B & 4 & 32 & 100 & 62 & -- \\
\midrule
Craftax & Codex & 3 & 8 & 0 & 0 & -- \\
 & Kimi-K2.6 & 1 & 1 & 100 & 0 & -- \\
\bottomrule
\end{tabular}
\end{table}

\textbf{The gap between the program's own checks and the general monitor appears for most author models, with small samples per model.} \label{app:detection}Table~\ref{tab:models} lists every model. Two entries need a comment. The Kimi-K2.6 programs differ widely from one another, so their interval covers the whole range. Half of the Mistral-Medium-3.5 programs in AppWorld carry a check that alarms on every day, changed or not, which accounts for all of that model's alarms. In Craftax the three Codex programs with checks produce eight failures, and their checks alarm on all eight; a single Kimi-K2.6 failure there raises no alarm.

\textbf{The program's own checks combine many misses with false alarms.} A battery is the set of evaluation episodes of a frozen program, ten in the original environment and ten under every change; the 20 reference episodes of the general monitor and the ten held-out ones are separate runs in that environment. On the unchanged episodes of the batteries the Codex checks alarm in 21\% of Seaquest episodes (800 episodes), in 10\% of Craftax episodes (30), and in no Kangaroo, service-autoscaling, or AppWorld episode (60, 40, and 40). Across the seven author models the service-autoscaling rate is 17\%, because five of the 18 programs carry a check that alarms in 30\% to 90\% of unchanged episodes. The general monitor alarms in 3\%, 2\%, and 0\% to 8\% (by author model) of unchanged episodes in these domains. The misses persist under stronger changes. In a graded version of the service-autoscaling changes, run on 35 programs from seven models, the misses of the own checks exceed those of the general monitor by at least 20 percentage points at all eight severity levels with enough failures, and when a capacity change makes all 35 programs fail, 61\% of the 18 with checks still raise no alarm.

\textbf{Monitors written on request are better than the program's own checks for the stronger models.} On the 168 Seaquest failures of programs with checks they miss 27\% when the author writes them and 33\% when a fresh session does; telling the writer to use normal records does not help (24\% and 39\%). On all 295 failures the four conditions miss 25\%, 27\%, 25\%, and 33\%, with false alarms on 1\% to 4\% of unchanged episodes. In service autoscaling they miss 0\% for Codex and 5\% for Claude Code, but 15\%, 17\%, 47\%, and 62\% for Qwen3.8-27B, gemma-4-31B, Kimi-K2.6, and Mistral-Medium-3.5.

\textbf{Simple progress detectors are not enough in Seaquest but are in service autoscaling.} On all 295 Seaquest failures, the general monitor misses 3\%, its success channel alone 35\%, and its structural statistics alone 6\%; a CUSUM and an EWMA detector on task progress miss 67\% and 46\%, and a learned anomaly detector on the same statistics misses 52\%. In service autoscaling every failure shows as a drop in the hours on target, and the success channel alone misses none of the 46 failures.

\textbf{Codex recognises failures from deployment records that it is handed afterwards; the open models mostly do not.} Each author session received eight unlabelled recorded deployment episodes of its own program, could analyse them freely within 60 tool calls, and had to say for each episode whether the environment had changed and whether the program was failing. In Seaquest, Codex answered ``changed'' for 120 of 120 episodes from failing changes, 0 of 80 unchanged episodes, and 91\% of episodes from harmless changes, and it judged the program to be failing in 94\% of the failing ones, in 9\% of those from harmless changes, and in none of the unchanged ones. In AppWorld it answered ``changed'' for 24 of 24 failing days and 0 of 16 unchanged days. The three open models, Qwen3.8-27B, Mistral-Medium-3.5, and gemma-4-31B, answered ``changed'' for 62\%, 50\%, and 10\% of their failing AppWorld days.

\begin{table}[h]
\caption{\textbf{Alarms with and without failure.} Share (\%) of failing units and of changed units with a working program on which a detector alarms (in at least half of the episodes), and share of unchanged episodes with an alarm, for Codex programs. The program's own checks are counted on the units of programs with checks (unit counts in brackets).}
\label{tab:alarms}
\centering
\small
\begin{tabular}{@{}llrrr@{}}
\toprule
Domain (failing / working units) & Detector & Failing units & Working units & Unchanged episodes \\
\midrule
Seaquest (295 / 705) & Program's own checks (168 / 332) & 48 & 23 & 21 \\
 & General monitor & 97 & 73 & 3 \\
 & \quad its structural statistics alone & 94 & 71 & 3 \\
 & \quad its success count alone & 65 & 10 & 1 \\
 & Monitor written on request by the author & 75 & 8 & 1 \\
\midrule
Kangaroo (34 / 56) & Program's own checks (22 / 38) & 45 & 8 & 0 \\
 & General monitor & 88 & 64 & 2 \\
\midrule
Service autoscaling (46 / 37) & Program's own checks (25 / 23) & 36 & 13 & 0 \\
 & General monitor & 100 & 51 & 0 \\
\midrule
AppWorld (58 / 38) & Program's own checks (28 / 20) & 11 & 0 & 0 \\
 & General monitor & 36 & 37 & 14 \\
\midrule
Craftax (9 / 51) & Program's own checks (8 / 28) & 100 & 32 & 10 \\
 & General monitor & 100 & 100 & 68 \\
\bottomrule
\end{tabular}
\end{table}

\textbf{A monitor written on request dominates the program's own checks on misses, false alarms, and alarms without failure.} Table~\ref{tab:alarms} gives, for each detector, the share of failing units and of changed units with a working program that are alarmed, and the share of unchanged episodes with an alarm. In Seaquest the own checks alarm on 48\% of failing units, 23\% of working units, and 21\% of unchanged episodes; the author's requested monitor on 75\%, 8\%, and 1\%. Of the 159 Seaquest units on which the own checks alarm, 81 are failures, about half.

\textbf{A rule on the outcome catches most failures where the outcome can be observed, and the checks written under the standard instructions do worse than this rule.} The rule alarms at the end of an episode whose success count (rescues, levels, hours on target, or achievements) is below half of the mean count of the program's 20 reference episodes; it was defined before it was computed, after the reviews of an earlier version asked for this baseline. On the failures of programs with checks it misses 7\% in Seaquest, 9\% in Kangaroo, 8\% and 5\% in service autoscaling (Codex and the six other models), and 25\% of the eight Craftax failures. It alarms on 11\%, 0\%, 0\%, 0\%, and 7\% of the changed units in which the program keeps working, and on 5\%, 3\%, 0\%, 2\%, and 0\% of unchanged episodes (Table~\ref{tab:detection}). Of the changed Seaquest units on which a detector alarms, 82\% are failures for this rule, 80\% for the monitor written on request, 51\% for the own checks, and 36\% for the general monitor. The rule needs a success signal that the program can read at run time, which AppWorld does not provide. Table~\ref{tab:changes-seaquest1} gives the counts change by change: the own checks miss 39\% of the 57 failures under the nine developer-written changes and 59\% of the 111 under the 16 blind ones, and five changes account for 187 of the 295 failures.

\textbf{Pooled over all standard programs with checks, the misses are half or more as point estimates, and the intervals are wide outside AppWorld.} Within each domain we pool every program with checks that was written under the standard instructions, whatever its author model and whichever study it belongs to, and resample programs and changes jointly. The own checks miss 63\% of 201 failures of 36 programs under the 16 blind Seaquest changes (95\% interval 40 to 83), 50\% of 26 failures of 7 programs in Kangaroo (8 to 100), 64\% of 107 failures of 18 programs in service autoscaling (44 to 83), and 84\% of 415 failures of 55 programs in AppWorld (73 to 93). The lower bound exceeds one half only in AppWorld. The general monitor misses 2\%, 19\%, 5\%, and 60\% of the same failures; the interval of the difference between the two excludes zero in Seaquest (38 to 82 points), includes zero in Kangaroo ($-$6 to 73 points), excludes zero in service autoscaling (39 to 79 points), includes zero in AppWorld ($-$8 to 59 points). The analysis was specified before it was computed, after a review asked for intervals that respect the small number of changes per domain.

\textbf{No threshold on the checks' verdicts gives a usable operating point.} Every stored episode holds, for each check, the number of steps on which it was confirmed and contradicted, so alarms can be recomputed under other rules; the behaviour of the program, which in one format skips rules when a check is broken, stays as recorded. In Seaquest (168 failures of 20 programs; the format's rule misses 52\% and alarms on 21\% of unchanged episodes), a rule that requires at least $k$ contradictions and more contradictions than confirmations misses 49\%, 52\%, 53\%, 61\%, 65\%, and 73\% for $k=1,2,3,5,10,20$, with alarms on 20\% of unchanged episodes at $k=1$ and 17\% at $k=20$. The recomputation uses the totals of an episode, whereas the engine applies its rule to the running counts, so the recomputed rule with $k=3$ (53\% missed, alarms on 18\% of unchanged episodes) is close to the format's rule but not identical with it. Without the comparison to confirmations the same counts miss 11\%, 19\%, 24\%, 35\%, 38\%, and 44\%, with alarms on 79\%, 66\%, 57\%, 53\%, 52\%, and 47\% of unchanged episodes. Only 4\% of the failing units show no contradiction in any episode: the checks are contradicted in normal play as well, and the counts do not separate the two. In service autoscaling (107 failures of 18 programs by seven models; the format's rule misses 64\%) the two families of rules miss 64\% to 92\% at alarm rates of 6\% to 9\%, and 27\% to 59\% at 38\% to 69\%. For the four Codex programs there, one contradiction that outnumbers the confirmations misses 44\% of 25 failures without any alarm on unchanged episodes, so a more sensitive threshold helps in this case and still leaves 44\% missed. In AppWorld the format already alarms on the first contradiction.

\textbf{The results do not depend on the failure threshold, and the failure set is not thinned by the checks.} With failure defined as keeping less than 0.3, 0.5, or 0.7 of the original performance, the own checks miss 51\%, 52\%, and 57\% of 129, 168, and 216 failures in Seaquest, 50\%, 55\%, and 52\% in Kangaroo, 73\%, 64\%, and 64\% for Codex in service autoscaling, and 89\% at each threshold in AppWorld; in Craftax they miss none of 4 and 8 failures and 27\% of 15. Between any two of the four sets of unchanged Seaquest episodes, no program drops below half of its performance (480 ordered pairs) and 1.5\% of the pairs drop below 0.7. Programs in the formats with checks fail more often under the first Seaquest catalogue than programs without, in 34\% against 25\% of units (difference 8 points, 95\% interval 4 to 12), at similar original scores (1.8 to 1.9 rescues per episode in each format). The format that only counts contradictions misses 58\% of its 78 failures, and the format that also skips the rules that rely on a broken check misses 47\% of 90, so we find no sign that checks which rescue a program remove failures from the set and thereby hide them from the comparison.

\textbf{A registered replication with new programs confirms the AppWorld result.} The first AppWorld study uses a failure criterion that was fixed after the two registered ones had given too few failures (Tables~\ref{tab:prereg} onwards). We therefore registered this criterion, a day succeeds when the whole report is correct and all 12 changes count as feasible, before 16 new Codex sessions wrote programs in the format with checks. The seven changes that turn out to cause failures are ones under which every failing program of the first study was repaired, so each has independently obtained successful executions. The registered prediction was at least 20 failures with the own checks silent on at least half of them and a lower interval bound above 30\%. The 16 programs fail under seven changes, 112 failures in all; their own checks miss 95\% (95\% interval over programs 91 to 98; over programs and changes 84 to 100) and the general monitor 62\%. The checks alarm on no original day and on none of the 80 changed units in which the program keeps working, where the general monitor alarms on 17\% and 40\%. Under the stricter feasibility rule of the second registered criterion, 32 failures remain, of which the own checks miss 88\% and the general monitor 50\%. The battery was run with a batched driver after the original one stalled, which changes process management only, not the programs, the episodes, or the scoring.

\subsection{Detection When the Agent Knows What the Checks Are For}

\textbf{New Seaquest sessions differ only in what the instructions say about the purpose of the checks.} \label{app:e4}Three conditions of 16 Codex sessions each use the format with checks and the same harness, with sessions interleaved across conditions. The first uses the standard instructions. The second adds one paragraph after the advice on writing assumptions: ``Your rules.py is then deployed: it plays on its own, episode after episode, and you are not there. During deployment the game may change in ways that nobody announces and that you cannot foresee: an action may stop having its usual effect or cost, or things may move or appear differently. Nobody watches the mission. The only thing that can call you back to repair the program is the program itself: an episode raises an alarm when one of its assumptions is broken in that episode, and an alarm calls you back. If the program stops achieving its mission and no assumption is broken, nobody is called and the program keeps failing. An alarm in an episode of the unchanged game calls you for nothing. So write the assumptions so that they are contradicted when the game no longer lets the program achieve its mission, whatever the reason, and are not contradicted in normal play.'' The third condition adds a tool that runs the current program with its checks for one or two episodes in one of five developer-written changes from the rounds before the first catalogue, labelled A to E without description, and reports the rescues and every check's verdicts; the instructions say that deployment will bring other changes. The alarm rule is unchanged. Each program is evaluated without the agent under the 16 blind changes of the first catalogue, and the general monitor uses 20 reference episodes of that program. One standard session ended after nine minutes with an infrastructure failure and no program; it was rerun once from scratch, as registered. One program of the third condition scores below the registered minimum in the original environment and is excluded. 

\begin{table}[h]
\caption{\textbf{Detection when the instructions state what the checks are for (informed).} New programs in the format with checks. Misses: share of failures (\%) without an alarm, with 95\% intervals from resampling programs and changes jointly (where every resample agrees, the exact binomial interval for the smaller of the numbers of programs and failing changes); difference: own checks minus general monitor on the same failures. Unchanged: share of unchanged episodes or days on which the own checks alarm. Working: share of changed units with a working program on which the own checks / the general monitor alarm. Progress: programs that declare an assumption on task progress; the last column gives the misses when such assumptions are left out. The progress columns apply to Seaquest only.}
\label{tab:informed}
\centering
\footnotesize
\setlength{\tabcolsep}{4pt}
\begin{tabular}{@{}lrrlrlrcrr@{}}
\toprule
Instructions & $P$ & $F$ & Own checks & General & Difference & Unchanged & Working & Progress & Without \\
\midrule
\multicolumn{10}{@{}l}{\emph{Seaquest, 16 blind changes}} \\
Standard & 16 & 90 & 69 [37, 93] & 1 & 68 [35, 93] & 6 & 11 / 70 & 0 & 69 \\
Informed & 16 & 84 & 6 [0, 21] & 4 & 2 [$-$9, 19] & 10 & 12 / 66 & 16 & 70 \\
Informed, test tool & 15 & 83 & 17 [3, 40] & 5 & 12 [$-$7, 38] & 11 & 11 / 60 & 15 & 72 \\
\midrule
\multicolumn{10}{@{}l}{\emph{AppWorld, 12 blind changes}} \\
Standard, Codex & 16 & 112 & 95 [84, 100] & 62 & 33 [0, 68] & 0 & 0 / 40 & -- & -- \\
Informed, Codex & 16 & 122 & 0 [0, 37] & 63 & $-$63 [$-$93, $-$29] & 0 & 9 / 23 & -- & -- \\
Standard, Qwen3.8-27B & 8 & 63 & 94 [76, 100] & 52 & 41 [$-$2, 82] & 0 & 0 / 24 & -- & -- \\
Informed, Qwen3.8-27B & 8 & 64 & 84 [61, 100] & 50 & 34 [$-$19, 83] & 0 & 0 / 22 & -- & -- \\
Standard, Mistral-Medium-3.5 & 8 & 61 & 51 [13, 87] & 61 & $-$10 [$-$58, 39] & 50 & 51 / 29 & -- & -- \\
Informed, Mistral-Medium-3.5 & 8 & 62 & 87 [61, 100] & 58 & 29 [$-$11, 69] & 13 & 12 / 29 & -- & -- \\
Standard, gemma-4-31B & 7 & 56 & 86 [57, 100] & 63 & 23 [$-$18, 67] & 14 & 14 / 25 & -- & -- \\
Informed, gemma-4-31B & 8 & 63 & 100 [63, 100] & 62 & 38 [4, 73] & 0 & 0 / 27 & -- & -- \\
\midrule
\multicolumn{10}{@{}l}{\emph{Seaquest, nine developer-written changes (the programs of the first two rows)}} \\
Standard & 16 & 50 & 38 [10, 92] & 6 & 32 [5, 79] & 6 & 11 / 69 & -- & -- \\
Informed & 16 & 50 & 4 [0, 21] & 4 & 0 [$-$20, 19] & 10 & 11 / 74 & -- & -- \\
\bottomrule
\end{tabular}
\end{table}

\textbf{Stating the purpose removes most misses, and both registered predictions fail.} Under the standard instructions the checks miss 69\% of 90 failures (95\% interval over programs and changes 37 to 93), in line with the 59\% of the first study (Table~\ref{tab:detection}; Table~\ref{tab:informed} gives the three conditions). With the purpose stated they miss 6\% of 84 (0 to 21), which is 2 percentage points more than the general monitor on the same failures (interval $-$9 to 19), and with the test tool 17\% of 83 (3 to 40; 12 points more than the general monitor, $-$7 to 38). We had registered that the informed checks would still miss at least 30\% and at least 20 points more than the general monitor, and that the checks of the third condition would miss at least 10 points more than the general monitor with an interval above zero; neither prediction holds. The registered expectation that the test tool would lower the misses further is not met either (11 points more than without it, interval 0 to 22). The informed checks alarm on 10\% and 11\% of unchanged episodes, against 6\% under the standard instructions, and on 12\% and 11\% of the changed units in which the program keeps working, where the general monitor alarms on 66\% and 60\%. Task performance does not differ (2.0, 2.1, and 2.2 rescues per episode). Sessions take 30, 29, and 48 minutes at the median, and every session of the third condition uses the test tool at least once (median 11 calls).

\textbf{The informed programs add checks on task progress, and their checks on mechanisms miss as much as before.} By our reading of the declared assumptions, every informed program (16 and 16 in the two conditions) contains at least one assumption that the mission progresses within a time budget, for example ``A six-diver delivery is observed within thirty-two measured initial-refill periods of the previous full delivery'', and none of the 16 standard programs does: theirs concern mechanisms such as movement, pickup, refill, unloading, and firing. When the alarms are recomputed from the stored verdicts without the progress assumptions, the informed programs miss 70\% and 72\% of their failures, as many as the standard programs; with the progress assumptions alone they miss 11\% and 18\%. Of the 79 failures that the checks of the second condition detect, 75 raise the alarm of a progress assumption and 54 of no other. Progress assumptions also raise most of the false alarms (9\% of unchanged episodes, against 1\% for the others). A progress assumption is a deadline that the author sets from quantities the program measures, such as the duration of a refill; it is stricter than the general monitor's floor on the success count, which alone misses 35\% of the failures of the first study. This breakdown was made after the results and was not registered. Stating the purpose thus changes what the checks cover, by adding a watchdog on the program's own progress, and leaves the checks on what the rules rely on as they were.

\textbf{In AppWorld, stating the purpose removes the misses as well, and the registered prediction fails.} The AppWorld instructions already state that the program is deployed, that nobody checks the report, and that an assumption that returns False raises the day's alarm. We add the paragraph used in Seaquest, adapted to the service: ``Your program.py is then deployed: it runs every day on its own, and you are not there. During deployment the service may change in ways that nobody announces and that you cannot foresee: an API may start to answer differently from how it answers now. Nobody checks the digest. The only thing that can call you back to repair the program is the program itself: a day raises an alarm when one of its assumptions returns False on that day, and an alarm calls you back. If the digest becomes wrong and no assumption returns False, nobody is called and the program keeps returning wrong digests. An alarm on a day of the unchanged service calls you for nothing. So write the assumptions so that they are contradicted when the service no longer lets the program return a correct digest, whatever the reason, and are not contradicted on normal days.'' Sixteen new Codex sessions with this paragraph are compared with the 16 sessions of the replication above, which ran on the same day under the standard instructions; the evaluation, the failure criterion, and the general monitor are those of the replication. Because an AppWorld program cannot observe whether its report is correct, we had registered that the informed checks would still miss at least half of the failures, with a lower interval bound above 30\%. The prediction fails. The 16 informed programs fail in 122 units under 8 changes, and their checks miss 0\% of these failures, against 95\% of 112 under the standard instructions; they alarm on 0\% of original days and on 9\% of the 70 changed units in which the program keeps working, all under one change that alters paging. The general monitor misses 63\% of the same failures (Table~\ref{tab:informed}). Under the stricter feasibility rule of the second criterion, 42 failures remain and the informed checks miss none. The informed programs are about three times as long as the standard ones (median 18,100 against 6,500 characters), their sessions use 43 tool calls at the median against 26, and their reports are correct on 96\% of original days against 100\%. By our reading, every informed program declares, next to checks on the form of the responses, an assumption that independent ways of reading the same information agree: that paged lists are complete and disjoint, that a filtered query returns what filtering the unfiltered list gives, and that list entries match the detail records. The standard programs check the fields and filters of the calls the report needs.

\textbf{The open models gain little from the same paragraph.} For each of three open models, eight AppWorld sessions with the paragraph and eight without it ran at the same time, with the evaluation of the Codex programs. The own checks miss, for Qwen3.8-27B: 84\% of 64 failures with the paragraph (61 to 100) against 94\% of 63 without it; Mistral-Medium-3.5: 87\% of 62 failures with the paragraph (61 to 100) against 51\% of 61 without it; gemma-4-31B: 100\% of 63 failures with the paragraph (63 to 100) against 86\% of 56 without it. They alarm on 0\%, 13\%, and 0\% of original days. The registered expectation was only that the paragraph would lower the misses for each model, which holds for Qwen3.8-27B and not for Mistral-Medium-3.5 and gemma-4-31B. A first attempt at the Qwen3.8-27B sessions was discarded before any program was read, because the gateway's request timeout cut off the longer replies of the informed sessions; both conditions were rerun with a longer timeout, which the other two models used from the start.

\textbf{Repair and a replay of deployments show that the informed checks call the agent in time.} Every failing unit of the 16 standard and the 16 informed Codex programs received one forced repair session without information, as in the first study (60 tool calls). The sessions repair 100\% of the 112 failures of the standard programs and 100\% of the 122 of the informed ones. A ten-day deployment is the ten evaluation days in a fixed order, with the service changing before day 4; a trigger opens at most one session at the end of the first day on which it alarms, the repaired program runs from the next day, and a session opened before the change is used up without effect. Value is the number of days with an entirely correct report relative to the original program on the original service. For the standard programs it is 0.33 without a call, 0.36 (0.34--0.38) with the own checks, 0.68 (0.64--0.72) with the general monitor, and 0.90 with a call on the day of the change; for the informed programs it is 0.32 (0.31--0.34), 0.90 (0.89--0.90), 0.64 (0.56--0.72), and 0.90 (0.89--0.90). The general monitor opens its session before the change in 0\% and 20\% of the deployments, the own checks in 0\% and 0\%; when alarms before the change are ignored, the general monitor reaches 0.68 and 0.72. The own checks of the informed programs give 0.54 more value than those of the standard programs (95\% interval 0.52 to 0.56). The replay covers failing deployments only, since changed units in which the program keeps working received no session.

\textbf{The informed checks were also evaluated under changes that no Codex session wrote.} The 16 standard and the 16 informed Seaquest programs were run under the nine developer-written changes of the first catalogue, which predate these sessions and were not used to write the paragraph, with the same seeds, alarm rule, and reference episodes. The informed checks miss 4\% of 50 failures (0 to 21) and the standard ones 38\% of 50 (10 to 92); the general monitor misses 4\% and 6\%. The registered prediction, at most 30\% missed by the informed checks, holds.

\subsection{Repair: Additional Results}

\textbf{The Seaquest repair set is dominated by three changes.} \label{app:repair}Its 107 failures come from 40 programs under seven feasible changes, and 102 of them from three changes, with 40, 36, and 26 failures (Table~\ref{tab:changes-seaquest2}). On the 104 failures on which the general monitor alarmed, which include 40, 36, and 23 under these three changes, programs keep at least 68\%, 94\%, and 73\% of their original performance after the call, up from 1\%, 2\%, and 14\%. Every one of the 107 failures received a session without any information about the change or the alarm. The comparison of what the call says was pre-registered as an equivalence test on the 104 failures on which the general monitor alarmed, since an alert presupposes an alarm: a one-bit alert is within 0.10 of a runtime report (difference $-0.01$, 90\% interval $-0.05$ to 0.02).

\textbf{The repairs address the change at hand.} Repaired AppWorld programs do no better under the other changes than before the repair (48\% against 47\% of days correct), and the agent repairs 27 of the 30 Seaquest failures on which the program's own checks had alarmed, as it does the failures on which they were silent.

\textbf{Repairs seldom harm the original behaviour for Codex and often do for the open models.} A sham call in the unchanged Seaquest environment leaves the mean score at 1.15 times its previous value, although four of 40 programs fall below 0.9; this gain is small next to the rise from 5\% to 81\% of performance kept under a change. Of the 58 Codex repair sessions in AppWorld, one produces a program that falls below half of its performance in the original environment; for Qwen3.8-27B, Mistral-Medium-3.5, and gemma-4-31B this happens in 14 of 64, 13 of 61, and 13 of 56 sessions.

\subsection{Coverage Intervention: Details}

\textbf{Informative relations are chosen with knowledge of the change; random relations and the full set are not.} \label{app:coverage}A relation is one statistic of one variable in the general monitor's vocabulary. It counts as informative for a change when it alarms for at least 30\% of the other programs under that change and as unrelated when it alarms for at most 2\%; a failing program's own records are never used to choose its relations. In Seaquest 6.9 informative relations are added per failure on average, and the unrelated condition draws the same number 50 times; 163 of the 168 failures have enough unrelated relations to draw from, and the three conditions are compared on those. Adding every relation, which needs no knowledge of the change, leaves 2\%, 4\%, and 15\% of failures missed in Seaquest, service autoscaling, and Kangaroo; the false-alarm rate, averaged over programs, rises from 16\% to 17\% in service autoscaling and from 1\% to 4\% in Kangaroo. With 1, 2, 4, 8, 16, and 32 random relations the missed share in Seaquest is 51\%, 47\%, 40\%, 25\%, 14\%, and 4\%. On the 16 blind Seaquest changes alone, the own checks, the informative relations, and the unrelated relations leave 59\%, 5\%, and 58\% of failures missed.

\begin{figure*}[h]
\centering
\includegraphics[width=\textwidth]{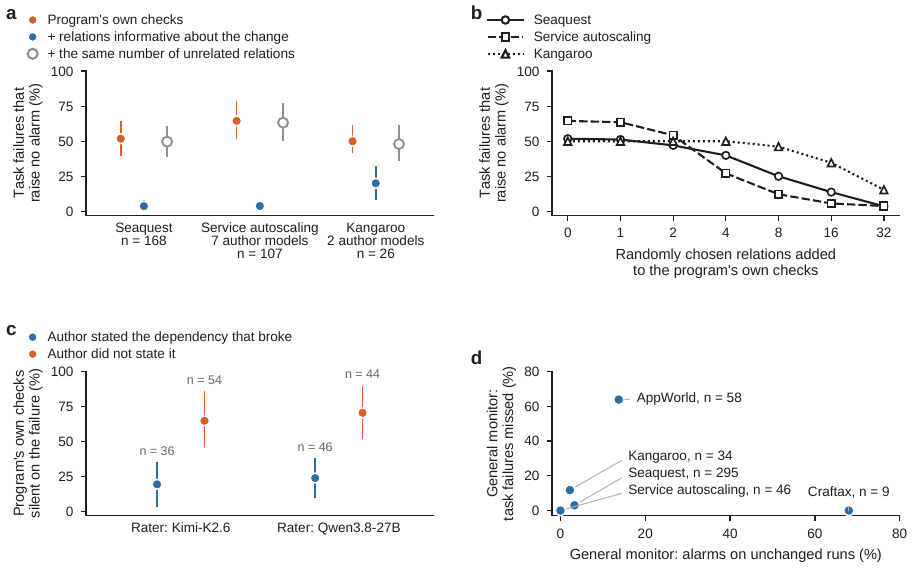}
\caption{\textbf{Coverage intervention, stated dependencies, and the general monitor in five domains.} (a) With the task program held fixed: its own checks; its own checks plus relations that are informative about the change; and its own checks plus as many unrelated relations. Service autoscaling pools the programs of all seven author models, and Kangaroo includes one Qwen3.8-Flash program. The two conditions with added relations are computed on the failures that have enough unrelated relations to draw from (163, 106, and 25), on which the own checks miss 52\%, 65\%, and 48\%. Alarm rates on unchanged runs are within about two points across the three conditions. (b) The same quantity as randomly chosen relations are added, which requires no knowledge of the change. (c) Failures on which the program's own checks are silent, by whether the author had stated the dependency that broke, as judged by two model raters blind to the detection outcomes; n counts program--change pairs of programs with declared assumptions. (d) The general monitor in five domains: failures missed against alarms on unchanged runs, over all failures. Whiskers in (a) and (c) are 95\% program-cluster bootstrap intervals.}
\label{fig:coverage-full}
\end{figure*}

\textbf{The raters agree moderately on which dependencies the authors had stated.} Cohen's kappa is 0.55 on whether the dependency was stated. Over all program formats the authors had stated the dependency that broke in 25\% and 41\% of the failing program--change pairs according to the two raters; among the 90 pairs of programs with declared assumptions shown in Figure~\ref{fig:coverage-full}c the shares are 40\% and 51\%. The comparison is observational, and stated and unstated dependencies may differ in the kind of change that breaks them, so we draw no causal conclusion.

\textbf{With the general monitor's reference episodes, the checks' contradictions still miss most failures, and the frequency with which the checks apply does not.} For each of the 20 Seaquest programs with checks we learn, from the 20 reference episodes, the range of every check's contradiction count per episode, widen it by a margin, keep the ranges that also hold in the ten held-out normal episodes, and alarm when a count leaves its range. With the registered margin this misses 41\% of the 168 failures and alarms on 21\% of unchanged episodes. Adding the range of every check's confirmation count misses 0.6\% and alarms on 28\%. A sweep of the margin, run after these results, traces the curves in Figure~\ref{fig:coverage}c: among the settings with alarms on at most 5\% of unchanged episodes, the contradictions alone miss 61\% (alarms on 3\%), and contradictions with confirmations miss 15\% (alarms on 3.9\%, and on 23\% of the changed units with a working program). The settings are chosen on the evaluation data, which favours the checks. A confirmation count measures how often a check applies, which the check itself never tests; in the example below it falls to zero. The setting with alarms on at most 5\% (a margin of one tenth of the range plus three, alarm when at least two counts leave their ranges) was then held fixed and applied to populations on which it was not chosen, each program with its own 20 reference and ten held-out episodes: the new Seaquest programs: 11\% of 90 failures missed (1 to 28), alarms on 5\% of unchanged episodes and 33\% of working units; Kangaroo: 18\% of 22 failures missed (0 to 55), alarms on 2\% of unchanged episodes and 26\% of working units; service autoscaling for Codex: 20\% of 25 failures missed (0 to 54), alarms on 0\% of unchanged episodes and 17\% of working units; the six other models there: 48\% of 82 failures missed (20 to 75), alarms on 0\% of unchanged episodes and 10\% of working units. For the informed Seaquest programs the same rule misses 18\% of 84 failures, against 6\% under the format's rule. The rule uses the same reference episodes as the general monitor and the outcome rule; the format's own rule uses none.

\subsection{A Worked Example of a Missed Failure}

\textbf{A check can be correct and never examine the relation that breaks.} \label{app:example}One Codex program for Seaquest contains a rule, \texttt{collect}, that pursues a visible diver. The rule declares three assumptions, on steering, on the oxygen cost of actions, and on pickup. The last one reads ``Overlapping a visible diver adds one carried diver while below capacity.'' Its check runs at every step, gives no verdict unless the submarine overlaps the diver it is pursuing, and in that case tests whether the number of carried divers rose. Under the blind change \texttt{evasive\_divers}, a diver within 40 pixels of the submarine in its lane turns to swim away from it. The program keeps pursuing, the submarine no longer overlaps a diver, and the check gives no verdict in any of the ten episodes; steering, firing, and oxygen behave as before, so the other assumptions hold. The program completes no rescue, against 3.6 per episode before the change, and its checks alarm in none of the ten episodes. What broke is that pursuit leads to an overlap, which the author neither stated nor checked. It is visible in the records: the rate at which the number of carried divers changes within a 500-step window leaves its normal range by step 1,500 of the first episode, and the general monitor alarms in all ten episodes under the change.

\subsection{Asking the Agent for Coverage: Details}

\textbf{Four instructions for a requested monitor give four operating points, none of them near the general monitor.} \label{app:asking}Each session is run by the Codex author of one of the 40 Seaquest programs after learning, with 60 tool calls. It can run the frozen program on up to ten unchanged episodes, analyse the recordings with its own code, and test a monitor on them. The monitors are scored on all 295 failures and on 1,600 unchanged episodes. A plain request misses 25\% of the failures (95\% interval 19 to 31) and alarms on 0.8\% of unchanged episodes; three times the budget at the highest reasoning effort gives 26\% and 1.5\%. An instruction to record every kind of object in the observations and the range of every number misses no failure and alarms on 99.8\% of unchanged episodes. The same instruction followed by a calibration requirement also misses no failure and alarms on 99.6\%. The requirement tells the author that ranges measured on a few episodes are too narrow, to keep only quantities whose ranges also hold on normal episodes kept aside, and to finish only when at least four such episodes raise no alarm. Every session's final test on its own recordings (six episodes at the median) raised no alarm, yet every monitor alarms on at least 90\% of the unchanged evaluation episodes of its program. The registered prediction that calibration would bring false alarms to 10\% of unchanged episodes or less failed.

\textbf{Examples of the program failing lower the misses under unseen changes by nine percentage points.} Each author received up to three recorded episodes of its own program, each under a different developer-written change that made the program fail, and was told that these were examples and that the monitor would be judged on other changes. Thirty of the 40 sessions analysed the examples and 23 tested their monitor on them. On the 199 failures under the 16 blind changes the monitors miss 19\% (11 to 27), against 28\% (20 to 36) for the plain request; the paired difference is 9.0 points (2.6 to 16.0), below the ten points that the pre-registered prediction required. They alarm on 1.3\% of unchanged episodes. On the 91 failures under changes of which the author had seen an episode they miss 4\%, against 18\% for the plain request, and on all 295 failures 14\% against 25\%. On the 111 blind failures of programs with checks, the own checks, the plain request, the request with examples, and the general monitor miss 59\%, 30\%, 24\%, and 3\%. Three of the 40 sessions ended after three to nine tool calls because the tool server failed to start, and we ran them again once from scratch. If these three count as never alarming, as the registered rule prescribes for a session that leaves no monitor, the missed share under blind changes is 24\% and the difference to the plain request is 4 points (95\% interval $-4$ to 12).

\subsection{Deployment Replay: Details and Sensitivity}

\textbf{The replay composes recorded outcomes day by day.} \label{app:replay}Each day is one evaluation episode. The program runs in the original environment before the change and in the changed one afterwards. A trigger opens a session after the first day on which it alarms, and from the next day the deployment uses the outcomes of the program that this session produced. A session opened before the change yields the program produced by a sham call. The perfect repairer replaces the outcomes after a session opened in the changed environment by those of the original program in the original environment.

\begin{figure*}[h]
\centering
\includegraphics[width=\textwidth]{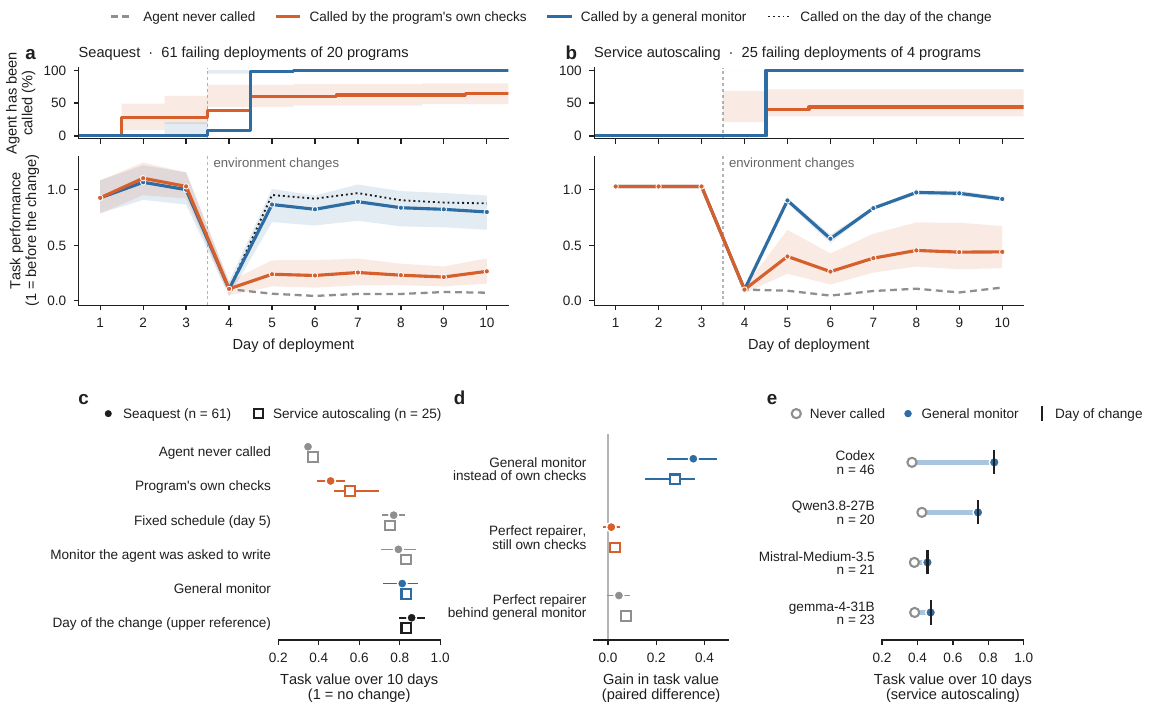}
\caption{\textbf{Deployment replay in both domains, paired gains, and results by author model.} (a, b) Replay of 10-day deployments in which the environment changes before day 4, for the failing deployments of programs that wrote their own checks. Top: share of deployments in which a repair session has been opened by the end of each day, when the session is triggered by the program's own checks or by the general monitor; sessions opened before day 4 are false alarms. Bottom: mean daily task performance relative to the performance before the change. At most one session is opened per deployment, and the same recorded repair sessions are replayed under every policy. (c) Ten-day task value by who calls the agent, on the same deployments. (d) Paired gains from replacing the program's own checks with the general monitor, and from replacing the real repairer with a perfect one, an offline upper bound, behind either trigger. (e) Service autoscaling, all failures, by author model. Bands and whiskers are 95\% program-cluster bootstrap intervals.}
\label{fig:value-full}
\end{figure*}

\textbf{The ranking of triggers is stable when the change comes earlier or later, except for the fixed schedule.} With the change before day 2, 4, 6, or 8, the general monitor yields 0.80, 0.81, 0.84, and 0.86 in Seaquest and 0.82, 0.83, 0.88, and 0.89 in service autoscaling on all failures; a call fixed on day 5 yields 0.58, 0.77, 0.57, and 0.78, and 0.56, 0.75, 0.54, and 0.71. Three scheduled calls on days 3, 6, and 9 yield 0.70 in service autoscaling. With the change before day 4, a call on a random day yields 0.56 and 0.55, and a delay of one, two, or three days after the general monitor's alarm costs 0.05, 0.14, and 0.21 of value in Seaquest. A rule that alarms when a day's success falls below half of the preceding days yields 0.82 in both domains, the same as the general monitor. AppWorld exposes no success signal at run time, so the rule cannot be used there.

\textbf{The loss from the program's own checks has two sources in Seaquest, and it remains when false alarms are free.} Of the 61 failing deployments of programs with checks, the checks never open a session in 22, open it before the change in 24, and after the change in 15. If alarms before the change neither use the session nor alter the program, the checks open a session after the change in 36 deployments and reach 0.60 (0.54--0.67), against 0.46 under the one-session rule and 0.81 for the general monitor; the difference to the general monitor is 0.21 (0.11--0.32). In service autoscaling the Codex checks raise no alarm before the change, so this bound equals the reported value. A perfect repairer behind the general monitor adds 0.04 in Seaquest and 0.07 in service autoscaling for Codex, and 0.44 and 0.41 for Mistral-Medium-3.5 and gemma-4-31B.

\textbf{The triggers differ in how often they open a session.} The replay opens a session on the first day with an alarm. This rule is more sensitive than the criterion for detection in the Evaluation Framework, which requires an alarm in at least half of the episodes, so the own checks open sessions in more deployments than they detect failures by that criterion. In Seaquest the own checks open a session in 64\% of the 61 failing deployments of programs with checks and the general monitor in 99\% of all 107 failing deployments; in service autoscaling the shares are 44\% and 100\%. A Codex repair session in service autoscaling takes 42 minutes and 32 tool calls on average, so the general monitor spends about 24 more minutes of session time per deployment than the own checks, for a gain of 0.28 in ten-day task value.

\textbf{With a cost per session and with deployments in which nothing fails, timely triggers keep most of their advantage.} In service autoscaling every changed unit of the Codex programs has a recorded session, so the replay can include the 37 units in which the program keeps working next to the 46 failing ones. On the working units the general monitor opens a session in 68\% of deployments, the requested monitor in 49\%, and the own checks in 39\% (of the 23 units of programs with checks); value is 0.93 without calls and 0.96, 0.97, and 0.94 with them, so these sessions do not harm the program. With a session charged at $\kappa$ days of full task value, net value is the value minus $\kappa$ times the number of sessions divided by ten. Over all 48 changed units of programs with checks the general monitor is ahead of the own checks by 0.16 at $\kappa=0$ and by 0.11 at $\kappa=1$ (95\% interval 0.07 to 0.16), and the two break even at $\kappa=3.7$; against never calling, the general monitor breaks even at $\kappa=3.2$ over all 83 changed units. On the working units alone its calls cost 0.04 of value at $\kappa=1$ (0.03 to 0.05). In a ten-day deployment without any change no trigger opens a session in service autoscaling; in Seaquest the own checks open one in 40\% of such deployments, the general monitor in 25\%, and the requested monitor in 5\%. Seaquest has recorded sessions only for failing units, where the general monitor is ahead of the own checks by 0.32 at $\kappa=1$ (0.21 to 0.42) and breaks even at $\kappa=9.8$, that is, when one session costs almost a whole deployment.

\textbf{A general-purpose language model that reads daily summaries is a poor trigger.} We gave GLM-5.2, which wrote none of the programs, the hourly summaries of ten normal reference days and of the deployment days so far, and asked after each day whether the service had changed; its first ``changed'' opened the session. The setting was fixed before the run (no extended reasoning, chosen by a rule on response time). On the 46 failing Codex deployments in service autoscaling it answered ``changed'' before the change in 57\% of them, and the resulting value is 0.55, which is 0.28 below the general monitor on the same deployments (0.14--0.42). On the 25 deployments of programs with checks it is 0.09 below the program's own checks, a difference that the interval does not separate from zero ($-0.33$ to 0.13). For the programs of Qwen3.8-27B, Mistral-Medium-3.5, and gemma-4-31B it answered ``changed'' before the change in 100\%, 71\%, and 100\% of deployments. Both pre-registered predictions for this trigger failed.

\subsection{Pre-Registered Predictions and Their Outcomes}

\textbf{Registrations are dated files in the project repository, and Tables~\ref{tab:prereg} onwards list every registered prediction with its outcome.} \label{app:prereg}Each registration file was written before the sessions or analyses it covers, and the job logs print its hash together with the hashes of the analysis code. The detection, repair, and replay comparisons of the main text were registered for each confirmation; analyses marked as exploratory or run after the results they follow are named as such where they are reported.

\begin{table}[h]
\scriptsize
\centering
\setlength{\tabcolsep}{4pt}
\caption{\textbf{Registered predictions and their outcomes.} Registrations are dated files in the project repository whose hashes are printed in the job logs, and this table and its three continuations list all registered predictions with their outcomes. ``Rule'' marks a registered reading rule without a predicted direction, for which ``held'' means that its threshold was met; ``misses'' is the share of failures without an alarm and ``false alarms'' the share of unchanged runs with one, ``diff.'' is the missed share of the general monitor minus that of the own checks, the opposite sign to the difference in Tables~\ref{tab:detection} and~\ref{tab:informed}, and CI is a 95\% interval unless noted. Codes in brackets are the identifiers of the registration files; Seaquest rounds are successive catalogues of changes, of which rounds 4 to 7 form the first catalogue of the paper.}
\label{tab:prereg}
\begin{tabular}{@{}p{0.24\textwidth}p{0.36\textwidth}p{0.24\textwidth}l@{}}
\toprule
Study & Registered prediction & Observed & Outcome \\
\midrule
\multicolumn{4}{@{}l}{\emph{Detection (Q1)}} \\
Seaquest round 1 (C1) & General monitor misses $\leq$ 0.10, and fewer than the own checks of one format (CI upper $<$ 0) & 21/50 = 0.42 [0.28, 0.56] & failed \\
Seaquest round 2 (C1) & As round 1, monitor revised & 13/64 = 0.20 [0.10, 0.32] & failed \\
Seaquest round 3 (C1) & Monitor revised again; needs $\geq$ 20 failures & 11 failures & undetermined \\
Seaquest round 4 (C1) & Monitor revised again & 4/76 = 0.053; diff.\ $-$0.13 [$-$0.36, 0.10] & failed \\
Seaquest round 5 (C1) & Same monitor as round 4 & 1/20 = 0.05; diff.\ $-$0.11 [$-$0.30, 0.00] & failed \\
Seaquest round 6 (C1) & Same monitor; own checks of both formats & 1/70 = 0.014; diff.\ $-$0.625 [$-$0.829, $-$0.435] & held \\
Seaquest round 7 (C1) & As round 6 & 3/129 = 0.023; diff.\ $-$0.521 [$-$0.662, $-$0.389] & held \\
Seaquest rounds 1--7 (C2) & General monitor: false alarms $\leq$ 0.05 & 0.022 to 0.048 in every round & held \\
Seaquest, Codex (M, P1) & Requested monitor misses $\geq$ 0.30 and $\geq$ 0.20 more than the general monitor & 0.26 [0.19, 0.33]; +0.23 [0.16, 0.30] & failed \\
Seaquest, Codex (M, P2) & Asking it to use recorded statistics lowers its misses (CI upper $<$ 0) & author 0.00 [$-$0.06, 0.06]; fresh session +0.06 [$-$0.04, 0.15] & failed \\
Seaquest, Codex (M, P3) & A fresh session's monitor is within $\pm$0.15 of the author's & +0.02 [$-$0.04, 0.09]; +0.08 [$-$0.01, 0.17] & held \\
Seaquest, monitors written by Kimi-K2.6 for the Codex programs (M, P1) & As Codex & 0.47 [0.37, 0.58]; +0.44 [0.36, 0.53] & held \\
Seaquest, Kimi-K2.6 monitors (M, P2) & As Codex & +0.07 [$-$0.07, 0.20]; $-$0.03 & failed \\
Seaquest, Kimi-K2.6 monitors (M, P3) & As Codex & +0.00 [$-$0.12, 0.13]; $-$0.10 [$-$0.21, 0.02] & held \\
Seaquest, Codex (J) & Rule: from records, the author calls $\geq$ 0.80 of failing and $\leq$ 0.20 of unchanged episodes changed & 120/120; 0/80 & held \\
Seaquest, Kimi (J) & Rule as for Codex & 22 of 40 sessions lost & undetermined \\
Kangaroo (C1-K) & As Seaquest round 1 & 4/34 = 0.118; diff.\ $-$0.36 [$-$0.47, $-$0.29] & failed \\
Kangaroo (C2-K) & As Seaquest C2 & 0.022 & held \\
Kangaroo (K, own checks) & Own checks miss $\geq$ 0.30 of failures & 0.55 & held \\
Autoscaling, Codex (C1-O) & As Seaquest round 1 & 0/46 = 0.00; diff.\ $-$0.64 [$-$0.79, $-$0.44] & held \\
Autoscaling, Codex (C2-O) & As Seaquest C2 & 0.00 & held \\
Autoscaling, Mistral (G1) & General monitor misses $\leq$ 0.10; false alarms $\leq$ 0.05 & misses 0.00; false alarms 0.075 & failed \\
Autoscaling, gemma (G1) & As above & misses 0.13; false alarms 0.05 & failed \\
Autoscaling, five others (G1) & As above & misses $\leq$ 0.09; false alarms $\leq$ 0.025 & held \\
Autoscaling, seven models (G2) & Own checks miss $\geq$ 0.30 of failures & 0.375 to 0.92 & held \\
Autoscaling, Qwen3.8-27B (G3) & Author judging records: thresholds as in J & 11/12; 0/8 & held \\
Autoscaling, Mistral, gemma (G3) & As above & 12/12; 2/8 for each & failed \\
Autoscaling, Kimi (G3) & As above & all 8 sessions lost & undetermined \\
Autoscaling, Kimi, Mistral (G4) & Requested monitor misses $\geq$ 0.10 more than the general monitor & +0.47 [0.22, 0.76]; +0.62 [0.24, 0.83] & held \\
Autoscaling, Qwen3.8-27B (G4) & As above & +0.10 [$-$0.10, 0.35] & held \\
Autoscaling, Codex, Claude Code, gemma (G4) & As above & 0.00; +0.05; +0.04 & failed \\
\bottomrule
\end{tabular}
\end{table}

\begin{table}[h]
\scriptsize
\centering
\setlength{\tabcolsep}{4pt}
\caption{\textbf{Registered predictions and their outcomes (continued).}}
\label{tab:prereg2}
\begin{tabular}{@{}p{0.24\textwidth}p{0.36\textwidth}p{0.24\textwidth}l@{}}
\toprule
Study & Registered prediction & Observed & Outcome \\
\midrule
\multicolumn{4}{@{}l}{\emph{Detection (Q1), continued}} \\
Craftax episodes (C1-C) & As Seaquest round 1; needs $\geq$ 20 failures & 15 failures; misses 0.27 [0, 0.53] & undetermined \\
Craftax episodes (C2-C) & As Seaquest C2 & 0.04 & held \\
Craftax (C, own checks) & Own checks miss $\geq$ 0.30 of failures & 0.10 & failed \\
Craftax 6000-step days (C1-C2) & As Seaquest round 1; needs $\geq$ 20 failures & 9 failures; misses 0 & undetermined \\
Craftax 6000-step days (C2-C2) & As Seaquest C2 & 0.68 [0.48, 0.84] & failed \\
AppWorld, Codex and three open models (C1-X9) & As Seaquest round 1; success counts correct report fields & no failures for any model & undetermined \\
AppWorld, four models (C2-X9) & As Seaquest C2, per day & 0.13 to 0.19 & failed \\
AppWorld, four models (C2-X9, secondary) & Same when an alarm needs two statistics out of range & 0 to 0.013 & held \\
AppWorld, Codex (X9b) & As Seaquest round 1; success = whole report correct (set after no failures were seen) & 18 failures; misses 0.44 & undetermined \\
AppWorld, three open models (X9b) & As above & misses 0.33 for each model & failed \\
AppWorld, four models (X9b, own checks) & Own checks miss $\geq$ 0.30 of failures & 0.45 to 1.00 & held \\
AppWorld, Codex (J-X9) & Rule as J, per day & 24/24; 0/16 & held \\
AppWorld, Qwen3.8-27B (J-X9) & Rule as J; between 0.50 and 0.80 is left open & 0.62 [0.50, 0.75]; 0 & undetermined \\
AppWorld, Mistral, gemma (J-X9) & Rule as J; $\leq$ 0.50 is read as not recognised & 0.50 [0.38, 0.67]; 0.10 [0, 0.24] & failed \\
\addlinespace
\multicolumn{4}{@{}l}{\emph{Repair (Q2)}} \\
Seaquest round 1 (C3) & Retention gain from alarm-triggered repair has CI lower $\geq$ 0.15 & +0.234 [0.130, 0.363] & failed \\
Seaquest round 2 (C3) & As round 1 & +0.071 [$-$0.015, 0.156] & failed \\
Seaquest round 3 (C3) & As round 1; needs $\geq$ 8 failing units & 1 unit & undetermined \\
Seaquest round 4 (C3) & As round 1 & +0.702 [0.629, 0.779] & held \\
Seaquest round 5 (C3) & As round 3 & 6 units; +0.320 [0.128, 0.546] & undetermined \\
Seaquest round 6 (C3) & As round 1; a repair is kept only if it scores higher & +0.738 [0.664, 0.813] & held \\
Seaquest round 7 (C3) & As round 6; the original program plays first & +0.366 [0.303, 0.428] & held \\
Seaquest rounds 1, 2, 4--7 (C4) & The monitor report in the repair prompt does not name the change & no disclosing terms & held \\
Seaquest round 6 (C5) & Retention back in the original environment has CI lower $\geq$ 0.80 & 0.861 [0.787, 0.923] & failed \\
Seaquest round 7 (C5) & As round 6 & 1.000 [1.000, 1.000] & held \\
Seaquest (A, H1) & One-bit alert within $\pm$0.10 of the runtime report (90\% CI) & $-$0.013 [$-$0.045, 0.019] & held \\
Seaquest (A, H2) & Retention gain after a one-bit alert has CI lower $\geq$ 0.15 & +0.74 [0.70, 0.79] & held \\
Seaquest (A, H4) & Rule: after a needless session the program keeps $\geq$ 0.90 of its score & 1.15 [1.06, 1.25]; 4 of 40 programs below 0.90 & held \\
Autoscaling, Codex (E1-O) & Missed by own checks and repaired when called: $\geq$ 0.30 of failures & 37/46 = 0.80 [0.70, 0.89] & held \\
Autoscaling, Qwen3.8-27B (E1-O) & As above & 0.65 [0.50, 0.80] & held \\
Autoscaling, Mistral, gemma (E1-O) & As above & 0.10 [0, 0.18]; 0.09 [0.08, 0.09] & failed \\
AppWorld, Codex (E1-X9) & Rule: the same quantity is $\geq$ 0.30 & 55/58 = 0.95 [0.89, 1.00] & held \\
AppWorld, Qwen3.8-27B, Mistral, gemma (E1-X9) & Rule as above & 0.98; 0.69; 0.77 & held \\
\bottomrule
\end{tabular}
\end{table}

\begin{table}[h]
\scriptsize
\centering
\setlength{\tabcolsep}{4pt}
\caption{\textbf{Registered predictions and their outcomes (continued).}}
\label{tab:prereg3}
\begin{tabular}{@{}p{0.24\textwidth}p{0.36\textwidth}p{0.24\textwidth}l@{}}
\toprule
Study & Registered prediction & Observed & Outcome \\
\midrule
\multicolumn{4}{@{}l}{\emph{Cause and coverage (Q3)}} \\
Seaquest (D1) & Authors had stated $\leq$ 0.50 of the dependencies that broke (two raters) & 0.25 [0.20, 0.31]; 0.41 [0.36, 0.46] & held \\
Seaquest (D2) & Stated share on failing pairs at most 0.20 above non-failing pairs & 0.25 vs 0.15; 0.41 vs 0.24 & held \\
Autoscaling (mechanism M1) & The general monitor's success-count floor alone misses $\leq$ 0.10 of failures & 0.00 to 0.05 for four authors; gemma 0.74 & held (4/5) \\
Autoscaling (mechanism M2) & Requested monitor misses more where the progress rule misses & gemma 0.24 vs 0.00; other groups $<$ 5 failures & held \\
Autoscaling and Kangaroo (mechanism M3) & Own checks miss more where the progress rule misses & Kangaroo +0.13 [$-$0.33, 0.42]; other groups $<$ 5 failures & undetermined \\
Seaquest (X1-a) & Adding related relations to the own checks cuts misses by $\geq$ 0.25, unrelated ones by $\leq$ 0.05; gap has CI lower $>$ 0.15 & 0.52 to 0.037 vs 0.50; gap 0.46 [0.35, 0.57] & held \\
Seaquest (X1-b) & Misses fall steadily as random relations are added & 0.51 to 0.036 for 1 to 32 relations & held \\
Autoscaling, seven models pooled (X1-M, a) & As X1-a & gap 0.59 [0.47, 0.73] & held \\
Autoscaling, gemma (X1-M, a) & As X1-a & unrelated relations also cut misses by 0.083 & failed \\
Autoscaling, Kimi (X1-M, a) & As X1-a & gap 0.39 [0.0, 1.0] & failed \\
Autoscaling, five others (X1-M, a) & As X1-a & gaps 0.42 to 0.83 & held \\
Kangaroo (X1-M, a) & As X1-a & gap 0.28 [0.23, 0.39] & held \\
Autoscaling and Kangaroo (X1-M, b) & As X1-b & 1.00 to 0.03; 1.00 to 0.24 & held \\
Seaquest (X2) & Classic progress detectors miss $\geq$ 0.25 of failures & 0.67 (CUSUM); 0.46 (EWMA) & held \\
Seaquest (X2) & A learned detector on the same features misses $\geq$ 0.05 more than the general monitor & 0.52 vs 0.031 & held \\
Seaquest (E2-1) & Asked for broad coverage, the requested monitor misses $\geq$ 0.10 less (CI upper $<$ 0) & 0.00 vs 0.25; false alarms 0.9975 & held \\
Seaquest (E2-2) & A tripled budget lowers its misses by less than 0.10 & +0.014 [$-$0.05, 0.08] & held \\
Seaquest (E2-3) & Broad coverage with tripled budget misses $\leq$ 0.10 & 0.00; false alarms 0.998 & held \\
Seaquest (E3-1) & With a calibration requirement: false alarms $\leq$ 0.10 and misses $\geq$ 0.10 below the plain request & false alarms 0.996 & failed \\
Seaquest (E3-2) & With failure examples: misses on blind changes $\geq$ 0.10 below the plain request (CI upper $<$ 0) & $-$0.040 [$-$0.121, 0.040]; $-$0.090 after rerun & failed \\
\bottomrule
\end{tabular}
\end{table}

\begin{table}[h]
\scriptsize
\centering
\setlength{\tabcolsep}{4pt}
\caption{\textbf{Registered predictions and their outcomes (continued).}}
\label{tab:prereg4}
\begin{tabular}{@{}p{0.24\textwidth}p{0.36\textwidth}p{0.24\textwidth}l@{}}
\toprule
Study & Registered prediction & Observed & Outcome \\
\midrule
\multicolumn{4}{@{}l}{\emph{Deployment value (Q4)}} \\
Seaquest (E5, Q1) & As trigger, the general monitor gains $\geq$ 0.10 in value over own checks (CI lower $>$ 0) & +0.35 [0.25, 0.45] & held \\
Seaquest (E5, Q2) & General monitor within $\pm$0.05 of a call on the day of the change (whole CI) & $-$0.047 [$-$0.094, $-$0.010] & failed \\
Seaquest (E5, Q3) & General monitor beats one scheduled call (CI lower $>$ 0) & +0.037 [$-$0.009, 0.075] & failed \\
Seaquest (E5, mechanism) & General monitor above the requested monitor & +0.033 [$-$0.025, 0.084] & undetermined \\
Autoscaling, Codex (E5, Q1) & As Seaquest Q1 & +0.28 [0.15, 0.36] & held \\
Autoscaling, Codex (E5, Q2) & As Seaquest Q2 & 0.000 [0, 0] & held \\
Autoscaling, Codex (E5, Q3) & As Seaquest Q3 & +0.082 [0.079, 0.085] & held \\
Autoscaling, Codex (E5, Q4) & Three scheduled calls gain $\leq$ 0.05 over the general monitor & $-$0.14 [$-$0.15, $-$0.13] & held \\
Autoscaling, Codex (E5, Q5) & Without a failure, the general monitor loses $\leq$ 0.05 against never calling & +0.030 [0.021, 0.039] & held \\
Autoscaling, Codex (E5, mechanism) & Requested monitor close to the general monitor & +0.001 [0, 0.003] & held \\
Autoscaling, Qwen3.8-27B (E5, Q1) & As Seaquest Q1 & +0.29 [0.24, 0.33] & held \\
Autoscaling, Mistral (E5, Q1) & As Seaquest Q1 & +0.099 [0.061, 0.144] & failed \\
Autoscaling, gemma (E5, Q1) & As Seaquest Q1 & +0.025 [0.022, 0.028] & failed \\
Autoscaling, same three models (E5, Q2) & As Seaquest Q2 & +0.001; 0.000; $-$0.002 & held \\
Autoscaling, same three models (E5, Q3) & As Seaquest Q3 & +0.058; +0.017; +0.020 & held \\
Autoscaling, Kimi (E5, Q1--Q3) & As Seaquest & all 44 sessions lost & undetermined \\
Seaquest (X7a) & Rule: general monitor with real repairer beats own checks with perfect repairer (CI above 0) & +0.34 [0.23, 0.45] & held \\
Autoscaling, Codex, Qwen3.8-27B, Mistral (X7a) & Rule as above & +0.25; +0.28; +0.10 & held \\
Autoscaling, gemma (X7a) & Rule as above & $-$0.02 [$-$0.07, 0.03] & undetermined \\
Autoscaling, Codex (X6a-1) & A language model as trigger is at most 0.05 behind the general monitor in value & 0.28 behind [0.14, 0.42] & failed \\
Autoscaling, Codex (X6a-2) & The language model calls before the change in $\leq$ 0.10 of deployments & 0.57 & failed \\
\addlinespace
\multicolumn{4}{@{}l}{\emph{Other studies}} \\
Seaquest (X3) & Fewer reference episodes (5, 10, 20) raise false alarms; misses change little & alarms 0.094, 0.063, 0.033; misses 0.085, 0.017, 0.031 & held \\
Seaquest and autoscaling (X3b) & A narrow reference raises false alarms by $\geq$ 0.02; misses within 0.05 & +0.043 and +0.17; $-$0.007 and $-$0.021 & held \\
Autoscaling, Codex (X4) & Rule: in each of three generations own checks miss $\geq$ 0.30 and the general monitor $\leq$ 0.10 & 0.95, 0.79, 0.64; 0.03, 0, 0 & held \\
Autoscaling (X10-1) & Stronger changes do not cause fewer failures & holds in all five families & held \\
Autoscaling (X10-2) & Own checks miss $\geq$ 0.20 more than the general monitor in every testable cell & 8 of 8 cells; +0.30 to +0.80 & held \\
Autoscaling (X10-4) & General monitor misses $\leq$ 0.10 when the CPU field hides the change & 0 at every level & held \\
Seaquest (E4-1) & Told that checks are the only alarm, own checks miss $\geq$ 0.30 and $\geq$ 0.20 above general monitor & 0.06; +0.02 [$-$0.05, 0.11] & failed \\
Seaquest (E4-2) & With detection feedback in development, own checks still miss $\geq$ 0.10 more than general monitor (CI lower $>$ 0) & +0.12 [0.00, 0.25] & failed \\
AppWorld (X9c-1) & New programs: $\geq$ 20 failures; own checks miss $\geq$ 0.50 (CI lower $>$ 0.30) & 112 failures; 0.95 [0.91, 0.98] & held \\
AppWorld (E4b-1) & Told that checks are the only call for repair, own checks still miss $\geq$ 0.50 (CI lower $>$ 0.30) & 122 failures; 0.00 & failed \\
AppWorld, Mistral-Medium-3.5 (E6-A) & With the paragraph, own checks miss less than without it & 0.87 against 0.51 & failed \\
AppWorld, gemma-4-31B (E6-A) & With the paragraph, own checks miss less than without it & 1.00 against 0.86 & failed \\
AppWorld, new programs (E6-B1) & Forced repair succeeds for $\geq$ 0.80 of the failures of standard and informed programs & 1.00; 1.00 & held \\
AppWorld, new programs (E6-B2) & Replay: informed own checks give $\geq$ 0.30 more value than standard own checks (CI lower $>$ 0) and are within 0.10 of a call on the day of the change & +0.54 [0.52, 0.56]; gap 0.00 & held \\
AppWorld, new programs (E6-B3) & Replay: standard own checks gain $\leq$ 0.10 over never calling & +0.03 & held \\
Seaquest (R16-1) & Informed programs under the nine developer-written changes: $\geq$ 12 failures, own checks miss $\leq$ 0.30 & 50 failures; 0.04 & held \\
\bottomrule
\end{tabular}
\end{table}

\textbf{Several pre-registered predictions failed, and we report them here.} The general monitor was predicted to miss at most 10\% of Kangaroo failures and missed 12\% (4 of all 34). Monitors written on request in Seaquest were predicted to miss at least 30\% and missed 26\% (author and fresh sessions pooled, all 295 failures), and the instruction to use normal records was predicted to help and did not. In the Seaquest replay the general monitor was predicted to come within 0.05 of a call on the day of the change and to beat the fixed schedule; its value was 0.05 lower than the former (95\% interval 0.01 to 0.09) and not distinguishable from the latter at the registered change day (difference 0.04, interval $-0.01$ to 0.08). Craftax was run in two designs. The first used the game's own episodes, whose median length of 211 to 250 steps is shorter than the general monitor's window of 500 steps; the five Codex programs gave 15 failures, below the registered minimum of 20, the general monitor missed 27\% of them, and the own checks missed 10\% of those from programs with checks, against a prediction of at least 30\%. The second design, registered afterwards and reported in the main text, deploys days of 6,000 steps in which the world restarts after a death: nine failures, eight of them from programs with checks, none missed by the own checks. There the general monitor was predicted to alarm on at most 5\% of unchanged days and alarmed on 68\%. In AppWorld the registered failure criterion, a halving of the number of correct fields, produced no failures. A second criterion, registered after the failure counts were known and before any alarm was examined, counts a day as failed unless the whole report is correct and a change as feasible only if some program gets a day entirely right. It excludes five changes and leaves 18 failures, below the registered minimum of 20; on these the general monitor misses 44\%, and the own checks miss 75\% of those from programs with checks. The exploratory set treats all 12 changes as feasible because each passes the recovery checks of the session that wrote it. The language-model trigger was predicted to come within 0.05 of the general monitor and to open at most 10\% of its sessions before the change; it was 0.28 lower and opened 57\% before the change. Both predictions about stating the purpose of the checks failed: informed Seaquest programs were predicted to miss at least 30\% of failures and missed 6\%, and with the test tool they were predicted to miss at least ten points more than the general monitor with an interval above zero and missed 12 points more (0 to 25). Informed AppWorld programs were predicted to miss at least half of their failures and missed 0\%. Two predictions about asking the agent for coverage failed: the calibration requirement was predicted to hold false alarms at or below 10\% and left them at 99.6\%, and failure examples were predicted to lower the missed share under blind changes by at least ten percentage points and lowered it by nine, or by four points when the three interrupted sessions count as never alarming, as the registered rule prescribes.

\subsection{Why Coverage Bounds Detection}

\textbf{Detection at a fixed false-alarm rate is limited by how distinguishable healthy and failing executions are in the monitor's evidence.} \label{app:bound}A deployed check usually observes a representation $z=\phi(h)$ of the execution history $h$: selected variables, counters, assertions, or recorded relationships. Suppose the distributions of $z$ under healthy and failing execution are $P_0$ and $P_1$. Any detector whose alarm depends only on $z$ then satisfies
\begin{equation}
\mathrm{TPR}\leq\mathrm{FPR}+\mathrm{TV}(P_1,P_0),
\label{eq:coverage}
\end{equation}
where $\mathrm{TV}$ is total variation distance. This standard testing bound follows directly by applying the definition of total variation to the event on which the detector alarms. In particular, identical observed distributions make healthy and failing executions indistinguishable to any detector restricted to that representation. The bound covers everything supplied to the monitor, including reference records. It does not establish which relations an agent omits; that is what the coverage intervention tests.

\subsection{Change Catalogues}

\textbf{Tables~\ref{tab:changes-seaquest1} to~\ref{tab:changes-craftax} list every change with the number of Codex programs that fail under it.} \label{app:catalogues}Blind changes were written by isolated sessions of Codex that received the environment code and a description of the task but neither the programs nor their results; the nine developer-written Seaquest changes predate them. A change counts as infeasible, and is left out before any repair session, when no frozen program completes the task under it in any episode; this removed one change of the second Seaquest catalogue and two Kangaroo changes. The rule can also remove changes that a repaired program would master, so we give a bound: if the removed Seaquest change were counted as 40 failures that are never repaired, the share of failures that are missed and repaired would be 47\% instead of 64\%. In AppWorld all 12 changes are kept: the session that wrote a change had to show a way to compute the report under it, and the repairs later gave a successful execution for every change that caused a failure, which confirms that the task remained feasible.

\begin{table}[h]
\caption{\textbf{Seaquest, first catalogue: the 25 changes used for detection.} Failing counts the frozen Codex programs that fail under the change, out of the 40 evaluated, and Checks restricts the count to the 20 programs that contain their own checks. The last three columns count failures without an alarm: Own, of the own checks among the failures of programs with checks; General and Outcome, of the general monitor and of the outcome rule among all failures under the change.}
\label{tab:changes-seaquest1}
\centering
\scriptsize\setlength{\tabcolsep}{4pt}
\begin{tabular}{@{}l p{0.34\textwidth} l r r r r r@{}}
\toprule
Change & What changes & Written by & Failing & Checks & Own & General & Outcome \\
\midrule
\texttt{controls\_swap\_y} & Underwater, the up and down controls are exchanged. & developer & 40/40 & 20/20 & 8 & 0 & 0 \\
\texttt{controls\_swap\_x} & Underwater, the left and right controls are exchanged. & developer & 28/40 & 16/20 & 4 & 1 & 4 \\
\texttt{lives\_one} & No spare lives: the life counter never exceeds one. & developer & 3/40 & 3/20 & 1 & 3 & 0 \\
\texttt{dive\_cost8} & Leaving the surface costs 8 units of oxygen. & developer & 3/40 & 3/20 & 3 & 0 & 0 \\
\texttt{wrap\_edges} & The left and right screen edges are joined for the submarine. & developer & 2/40 & 1/20 & 1 & 0 & 0 \\
\texttt{emissile\_homing} & Enemy missiles in flight move one pixel per step toward the submarine's depth. & developer & 12/40 & 8/20 & 2 & 0 & 3 \\
\texttt{pickup\_cost10} & Picking up a diver costs 10 units of oxygen. & developer & 3/40 & 3/20 & 1 & 0 & 2 \\
\texttt{pickup\_gift10} & Picking up a diver gives 10 units of oxygen. & developer & 0/40 & 0/20 & 0 & 0 & 0 \\
\texttt{load\_slows} & Underwater with three or more divers aboard, the submarine moves only every second step. & developer & 5/40 & 3/20 & 2 & 1 & 1 \\
\addlinespace
\texttt{launch\_recoil} & Launching a torpedo underwater pushes the submarine six pixels opposite the shot. & blind session & 4/40 & 3/20 & 0 & 0 & 1 \\
\texttt{diver\_vertical\_routes} & Divers follow synchronized vertical routes up to 16 pixels above their original depths. & blind session & 4/40 & 4/20 & 2 & 0 & 0 \\
\texttt{friendly\_fire} & A player torpedo destroys an uncollected diver on overlap and is consumed. & blind session & 40/40 & 20/20 & 18 & 0 & 0 \\
\texttt{piercing\_torpedo} & A player torpedo survives an enemy hit and keeps flying, allowing further kills. & blind session & 1/40 & 1/20 & 0 & 1 & 1 \\
\texttt{snorkel\_refill} & Oxygen also refills in a band just below the surface, without surfacing. & blind session & 1/40 & 1/20 & 1 & 0 & 0 \\
\texttt{submarine\_waves} & Every shark is replaced by a missile-capable enemy submarine. & blind session & 0/40 & 0/20 & 0 & 0 & 0 \\
\texttt{diagonal\_priority} & An underwater diagonal move keeps its horizontal part and loses its vertical part. & blind session & 20/40 & 11/20 & 5 & 0 & 4 \\
\texttt{combat\_bounty} & Each torpedo kill awards 1,000 points instead of 20--90. & blind session & 0/40 & 0/20 & 0 & 0 & 0 \\
\texttt{evasive\_divers} & Divers near the submarine turn horizontally away from it. & blind session & 40/40 & 20/20 & 16 & 1 & 0 \\
\texttt{inward\_enemy\_spawns} & New sharks and enemy submarines enter inside the screen instead of at an edge. & blind session & 3/40 & 2/20 & 0 & 0 & 0 \\
\texttt{turn\_guided\_torpedo} & A torpedo in flight takes the submarine's current facing as its heading. & blind session & 7/40 & 4/20 & 2 & 1 & 1 \\
\texttt{triple\_waves} & Every new underwater enemy wave has three members. & blind session & 15/40 & 10/20 & 6 & 0 & 1 \\
\texttt{oxygen\_salvage} & Each underwater torpedo kill restores 12 units of oxygen, up to the usual maximum. & blind session & 0/40 & 0/20 & 0 & 0 & 0 \\
\texttt{blast\_radius} & A torpedoed enemy explodes and destroys other enemies within 36 pixels. & blind session & 0/40 & 0/20 & 0 & 0 & 0 \\
\texttt{stern\_docking} & A diver boards only within three horizontal pixels of the submarine's stern. & blind session & 39/40 & 20/20 & 8 & 0 & 0 \\
\texttt{death\_volley} & Each lane with a torpedoed enemy fires one revenge missile toward the submarine. & blind session & 25/40 & 15/20 & 7 & 1 & 3 \\
\bottomrule
\end{tabular}
\end{table}

\begin{table}[h]
\caption{\textbf{Seaquest, second catalogue: the eight blind changes used for repair and replay.} Failing counts the frozen Codex programs that fail under the change, out of the 40 evaluated; $^{\dagger}$\texttt{shark\_predation} was judged infeasible (no program completed a full rescue under it in any episode) and was therefore excluded before any repair session.}
\label{tab:changes-seaquest2}
\centering
\small
\begin{tabular}{@{}l p{0.46\textwidth} l r@{}}
\toprule
Change & What changes & Written by & Failing \\
\midrule
\texttt{central\_ascent\_hatch} & The submarine can ascend to the surface only within a central horizontal band. & blind session & 36/40 \\
\texttt{launch\_jettisons\_passenger} & Each underwater torpedo launch loses one carried diver. & blind session & 40/40 \\
\texttt{shark\_predation}$^{\dagger}$ & A shark eats any uncollected diver at its depth within eight horizontal pixels. & blind session & --- \\
\texttt{enemy\_forward\_shields} & An enemy absorbs a head-on torpedo within 24 pixels ahead of its front edge. & blind session & 26/40 \\
\texttt{turning\_brake} & Reversing horizontal facing underwater spends the step turning in place. & blind session & 2/40 \\
\texttt{paired\_passengers} & Each diver pickup adds two to the carried count, capped at six. & blind session & 0/40 \\
\texttt{boarding\_life\_award} & Each diver pickup awards a spare life, up to the usual six. & blind session & 1/40 \\
\texttt{combat\_escalation} & Each torpedo kill advances the difficulty by two levels, up to level seven. & blind session & 2/40 \\
\bottomrule
\end{tabular}
\end{table}

\begin{table}[h]
\caption{\textbf{Kangaroo: the 12 blind changes.} Failing counts the Codex programs that fail under the change, out of the nine evaluated; $^{\dagger}$two changes were judged infeasible (no program completed a level under them in any episode) and were excluded.}
\label{tab:changes-kangaroo}
\centering
\small
\begin{tabular}{@{}l p{0.52\textwidth} l r@{}}
\toprule
Change & What changes & Written by & Failing \\
\midrule
\texttt{ground\_inertia} & The kangaroo coasts when input is released and needs eight steps to reverse; crouching brakes. & blind session & 1/9 \\
\texttt{ladder\_rest\_cycle} & Climbing pauses for 32 of every 96 steps while the game continues. & blind session & 7/9 \\
\texttt{platform\_recovery} & Finishing a jump or climb blocks the next one for 90 steps. & blind session & 6/9 \\
\texttt{auto\_hop} & Fifteen consecutive walking steps trigger an automatic jump. & blind session & 7/9 \\
\texttt{drifting\_fruit} & Fruit sweeps horizontally up to 24 pixels either side of its original position. & blind session & 0/9 \\
\texttt{bell\_elevator} & The bell moves to the ground floor; activating it lifts the kangaroo one platform. & blind session & 5/9 \\
\texttt{pursuing\_monkeys} & Walking monkeys pursue the kangaroo's horizontal position at their original speed. & blind session & 2/9 \\
\texttt{homing\_drop} & The falling coconut keeps steering toward the kangaroo, one pixel every eight steps. & blind session & 6/9 \\
\texttt{sinking\_coconuts} & Thrown coconuts sink one pixel every four steps of flight. & blind session & 0/9 \\
\texttt{conveyor\_floors}$^{\dagger}$ & Floors carry a standing kangaroo one pixel every six steps; crouching anchors it. & blind session & --- \\
\texttt{wide\_player}$^{\dagger}$ & The kangaroo's physical width doubles from eight to sixteen pixels. & blind session & --- \\
\texttt{instant\_throws} & Thrown coconuts start flying immediately, skipping the stationary windup. & blind session & 0/9 \\
\bottomrule
\end{tabular}
\end{table}

\begin{table}[h]
\caption{\textbf{Service autoscaling: the 12 blind changes.} Failing counts the Codex programs that fail under the change, out of the seven evaluated; $^{\dagger}$one program stopped with errors under \texttt{cpu\_headroom}, a failure that announces itself, and is left out of the count.}
\label{tab:changes-autoscaling}
\centering
\small
\begin{tabular}{@{}l p{0.50\textwidth} l r@{}}
\toprule
Change & What changes & Written by & Failing \\
\midrule
\texttt{slow\_boot} & New servers take fifteen minutes to boot instead of three. & blind session & 2/7 \\
\texttt{compact\_workers} & Each running server handles 35 requests per step instead of 50. & blind session & 0/7 \\
\texttt{square\_wave\_load} & Traffic alternates between 60\% and 140\% of the daily load every five minutes. & blind session & 5/7 \\
\texttt{cpu\_headroom}$^{\dagger}$ & The CPU field reports percent idle capacity instead of percent busy. & blind session & 5/6 \\
\texttt{high\_utilization\_hours} & A good hour requires 70\% mean utilization instead of 50\%. & blind session & 7/7 \\
\texttt{saturation\_latency} & CPU contention adds latency for utilization above 60\%, even without a queue. & blind session & 7/7 \\
\texttt{removal\_handoff} & Removing a server adds 100 unfinished requests to the queue. & blind session & 7/7 \\
\texttt{overload\_retries} & Outstanding and dropped requests are retried next step at 35\%, capped at 400. & blind session & 0/7 \\
\texttt{paired\_provisioning} & Each requested addition provisions twice as many servers, up to the fleet cap. & blind session & 0/7 \\
\texttt{partitioned\_service} & Two isolated tenants split capacity equally but receive 70\% and 30\% of traffic. & blind session & 7/7 \\
\texttt{delayed\_telemetry} & Service measurements arrive two minutes late; fleet and accounting fields stay current. & blind session & 6/7 \\
\texttt{extended\_breach\_grace} & Only hours with more than 90 SLO violations spend error budget, instead of 18. & blind session & 0/7 \\
\bottomrule
\end{tabular}
\end{table}

\begin{table}[h]
\caption{\textbf{AppWorld: the 12 blind changes.} Failing counts the Codex programs that fail under the change, out of the eight evaluated; the last three changes were designated harmless controls by the session that wrote them.}
\label{tab:changes-appworld}
\centering
\small
\begin{tabular}{@{}l p{0.46\textwidth} l r@{}}
\toprule
Change & What changes & Written by & Failing \\
\midrule
\texttt{liked\_song\_offsets} & The Spotify liked-songs page index is a row offset instead of a page number. & blind session & 2/8 \\
\texttt{transactions\_oldest\_first} & Venmo transaction lists default to oldest first when no sort is requested. & blind session & 0/8 \\
\texttt{transactions\_default\_sent} & Venmo transaction lists default to sent transactions instead of both directions. & blind session & 8/8 \\
\texttt{liked\_dates\_day\_first} & Spotify liked-song dates use DD/MM/YYYY, keeping the time suffix. & blind session & 8/8 \\
\texttt{transaction\_like\_origin} & Venmo transaction lists report each like count plus one. & blind session & 8/8 \\
\texttt{playlist\_track\_previews} & The Spotify playlist library lists only the first track of each playlist. & blind session & 8/8 \\
\texttt{transaction\_email\_escaping} & Venmo transaction lists percent-encode the local part of each email address. & blind session & 8/8 \\
\texttt{incoming\_means\_one\_way} & The Gmail inbox list marks a thread incoming only if it has no outgoing mail. & blind session & 8/8 \\
\texttt{transaction\_party\_roles} & Venmo transaction lists swap the sender and the receiver. & blind session & 8/8 \\
\addlinespace
\texttt{description\_line\_endings} & Venmo transaction descriptions gain one trailing newline (control). & blind session & 0/8 \\
\texttt{thread\_key\_order} & Gmail inbox-thread records list their keys in reverse order (control). & blind session & 0/8 \\
\texttt{artist\_credit\_order} & Spotify liked-song records list artist credits in reverse order (control). & blind session & 0/8 \\
\bottomrule
\end{tabular}
\end{table}

\begin{table}[h]
\caption{\textbf{Craftax: the 12 blind changes.} Failing counts the Codex programs that fail under the change, out of the five evaluated.}
\label{tab:changes-craftax}
\centering
\small
\begin{tabular}{@{}l p{0.52\textwidth} l r@{}}
\toprule
Change & What changes & Written by & Failing \\
\midrule
\texttt{pivot\_steps} & Turning consumes a movement action; only a repeated heading walks. & blind session & 0/5 \\
\texttt{two\_hit\_trees} & Felling a tree requires two consecutive \texttt{do} actions facing it. & blind session & 3/5 \\
\texttt{six\_wood\_tables} & Placing a crafting table costs six wood instead of two. & blind session & 3/5 \\
\texttt{brittle\_pickaxes} & Each pickaxe breaks after eight successful mining uses of its tier. & blind session & 0/5 \\
\texttt{exertion\_thirst} & Each successful walking step adds three extra units to the hidden thirst timer. & blind session & 0/5 \\
\texttt{furnace\_recovery} & Healing progress is discarded after every step without a neighbouring furnace. & blind session & 0/5 \\
\texttt{path\_beds} & Sleep can start only while standing on a path tile. & blind session & 0/5 \\
\texttt{flash\_crops} & Planted crops mature in twenty steps instead of six hundred. & blind session & 0/5 \\
\texttt{armoured\_cows} & Cows have nine health instead of three. & blind session & 0/5 \\
\texttt{daylight\_truce} & While daylight is at least 50\%, zombies still chase but their attack cooldown stays at least two. & blind session & 0/5 \\
\texttt{deficit\_bars} & Health, food, drink and energy show missing points: zero is full, nine is empty. & blind session & 1/5 \\
\texttt{delayed\_view} & Creatures, terrain and facing are shown as of six steps ago; numbers stay current. & blind session & 2/5 \\
\bottomrule
\end{tabular}
\end{table}

\subsection{Extended Related Work}

\textbf{Software testing, dynamic invariant detection, and runtime verification provide foundations for determining whether program execution remains correct.} \label{app:related}The oracle problem concerns how to determine whether an execution is correct \citep{barr2015oracle}. Dynamic invariant detection infers likely properties from observed executions \citep{ernst2007daikon}, runtime verification checks executions against stated properties while a system runs \citep{leucker2009rv}, and ClearView uses learned invariants to detect and patch errors in deployed software \citep{perkins2009clearview}. Autonomic computing proposed systems that monitor, analyse, plan, and execute their own maintenance \citep{kephart2003autonomic}. Independent implementations need not fail independently \citep{knight1986independence}, which is why we separate who writes a check from what evidence it uses when comparing detectors on the same failures.

\textbf{Monitoring deployed learning systems for change is a long-standing problem.} Concept-drift and dataset-shift detection compare deployment data with reference data \citep{gama2014drift,rabanser2019failing}, with sequential tests such as CUSUM and exponentially weighted averages as classical tools \citep{page1954cusum,roberts1959ewma}. Data validation and model assertions apply declared or learned constraints to machine-learning pipelines \citep{breck2019datavalidation,schelter2018deequ,kang2020assertions}, and production guidelines list monitoring as a requirement that test accuracy does not cover \citep{sculley2015debt,breck2017mltest}. For agents, novelty in open worlds has been formalised and benchmarked \citep{boult2021novelty,balloch2022novgrid}, controlled modifications of Atari games test robustness \citep{delfosse2024hackatari}, and policies can pursue the wrong goal after a shift while remaining competent \citep{langosco2022goal}.

\textbf{Feedback-driven self-improvement is measured by task performance.} Iterative refinement and verbal feedback can improve language-model and agent outputs \citep{madaan2023selfrefine,shinn2023reflexion}, and work on intrinsic self-correction shows why the source of useful information must be controlled \citep{huang2024selfcorrect}. Agents that write programs as policies or skills verify them against task feedback \citep{liang2023cap,wang2023voyager}, and program search improves programs by their scores \citep{romera2024funsearch,hu2025adas}. The term recursive self-improvement is used for several forms of feedback-driven development: RSIGame tests and revises generated games \citep{wu2026rsigame}, SO-RSI diagnoses execution failures to guide agent-workflow revisions \citep{jiang2026sorsi}, STOP applies a program improver to itself \citep{zelikman2024stop}, and the Darwin G\"odel Machine modifies a coding agent's own implementation \citep{zhang2025dgm}.

\textbf{Whether models can state what their code relies on, and when agents should ask for help, are studied apart from deployment.} Language models can infer invariants and postconditions for given programs \citep{pei2023invariants,endres2024nl2postcond}, planners can be calibrated to ask for help under uncertainty \citep{ren2023knowno}, and failures of multi-agent systems can be attributed to agents and steps after the fact \citep{zhang2025whowhen,cemri2025mast}. Runtime monitoring and recovery for agents connect execution evidence to opportunities for intervention: telemetry-based detection combined with rollback and re-execution \citep{dubey2026realtime}, comparison of unfolding trajectories with successful executions \citep{barazandeh2026ontrack}, lessons retained around verified recoveries \citep{ji2026sentry}, and simple structural baselines that show what a monitor actually learns \citep{naagarajan2026structural}. These works build external monitors or study isolated capabilities. We measure, on the same failures, what the agent's own checks detect, what the agent could repair, and what the difference costs. Deciding when to spend computation on investigation is an instance of metareasoning \citep{russell1991metareasoning}.

\subsection{Figure Evidence and Reproducibility}

\textbf{Each analysis uses one of the failure populations in Table~\ref{tab:populations}.} \label{app:repro}Detection in Table~\ref{tab:detection}, the coverage test, and the requested monitors use the first Seaquest catalogue; repair and the deployment replay use the second, on which detection is measured as well. Analyses of the program's own checks are restricted to programs with checks, and the general monitor and the requested monitors are also reported on all failures. The coverage test in Kangaroo adds one Qwen3.8-Flash program with checks, which gives 26 failures in place of the 22 of the Codex programs.

\begin{table}[h]
\caption{\textbf{Failure populations.} Programs and task failures in each domain and change catalogue, for all programs and for the programs that contain their own checks; the lower part lists the programs written for the later experiments.}
\label{tab:populations}
\centering
\footnotesize
\begin{tabular}{@{}lrrrr@{}}
\toprule
Population & Programs & Failures & Programs with checks & Their failures \\
\midrule
Seaquest, first catalogue (25 changes), Codex & 40 & 295 & 20 & 168 \\
\quad of which under the 16 blind changes & 40 & 199 & 20 & 111 \\
Seaquest, second catalogue (7 blind changes), Codex & 40 & 107 & 20 & 61 \\
Kangaroo, Codex & 9 & 34 & 6 & 22 \\
Service autoscaling, Codex & 7 & 46 & 4 & 25 \\
Service autoscaling, seven author models & 35 & 188 & 18 & 107 \\
AppWorld (exploratory set), Codex & 8 & 58 & 4 & 28 \\
AppWorld (exploratory set), three open models & 23 & 181 & 12 & 95 \\
Craftax, Codex & 5 & 9 & 3 & 8 \\
Kangaroo, Qwen3.8-Flash & 2 & 8 & 1 & 4 \\
Craftax, Kimi-K2.6 & 4 & 4 & 1 & 1 \\
\midrule
\multicolumn{5}{@{}l}{\emph{New programs, all in the format with checks}} \\
Seaquest, 16 blind changes, Codex: standard instructions & 16 & 90 & 16 & 90 \\
\quad informed & 16 & 84 & 16 & 84 \\
\quad informed, with the test tool & 15 & 83 & 15 & 83 \\
Seaquest, 9 developer-written changes, Codex: standard & 16 & 50 & 16 & 50 \\
Seaquest, 9 developer-written changes, Codex: informed & 16 & 50 & 16 & 50 \\
AppWorld (registered replication), Codex: standard & 16 & 112 & 16 & 112 \\
\quad informed & 16 & 122 & 16 & 122 \\
AppWorld, Qwen3.8-27B: standard & 8 & 63 & 8 & 63 \\
\quad informed & 8 & 64 & 8 & 64 \\
AppWorld, Mistral-Medium-3.5: standard & 8 & 61 & 8 & 61 \\
\quad informed & 8 & 62 & 8 & 62 \\
AppWorld, gemma-4-31B: standard & 7 & 56 & 7 & 56 \\
\quad informed & 8 & 63 & 8 & 63 \\
\bottomrule
\end{tabular}
\end{table}

\textbf{Measured panels in Figures~\ref{fig:uninvoked}--\ref{fig:value} are drawn from frozen result files with comparison-specific failure populations.} Figure~\ref{fig:uninvoked}a classifies every failure and counts programs without checks as raising no alarm; its AppWorld rows use the exploratory failure set. Figure~\ref{fig:uninvoked}b uses the failures of Codex programs that wrote checks, so that all displayed monitors share a denominator. Figure~\ref{fig:coverage}a--b uses the failures of all programs with checks that have a coverage result: Codex in Seaquest, all seven author models in service autoscaling, and Codex with one Qwen3.8-Flash program in Kangaroo. Figure~\ref{fig:value}a--b and Figure~\ref{fig:value-full}a--c are restricted to the failing deployments of programs with checks so that every trigger is compared on the same deployments, Figure~\ref{fig:value}c adds the deployments in which these programs keep working, and Figure~\ref{fig:value-full}d treats the perfect repairer as an offline upper bound. Figures~\ref{fig:coverage-full} and~\ref{fig:value-full} repeat the main panels together with the panels that are not in the main text; the axis labels of Figure~\ref{fig:coverage-full}a count all failures of programs with checks (168, 107, and 26), whereas the two conditions with added relations use the 163, 106, and 25 failures named in its caption. Figures~\ref{fig:loop} and~\ref{fig:framework} are schematic.

\textbf{Figure documentation separates schematic diagrams from measured evidence.} The empirical figures show 95\% program-cluster bootstrap intervals, taken from the source analyses where these report them. Displayed descriptive rates without intervals are not population guarantees. In particular, the zero observed misses in service autoscaling do not establish zero risk under new changes. The figure scripts read the frozen result files, recompute the closed-loop quantities with the frozen replay code, and export vector PDF and SVG files with PNG previews and the plotted values.

\end{document}